\documentclass[final,3p,times,twocolumn]{elsarticle}
\usepackage{amssymb}
\usepackage{array}
\usepackage[caption=false,font=normalsize,labelfont=sf,textfont=sf]{subfig}
\usepackage{textcomp}
\usepackage{stfloats}
\usepackage{subfig}  %插入多图时用子图显示的宏包
\usepackage{subcaption} % 如果使用 \subfigure
\usepackage{amsmath}
\usepackage{hyperref}
\usepackage{cleveref}
\usepackage{verbatim}
\usepackage{graphicx}
\usepackage{balance}
\usepackage{multirow}
\usepackage{booktabs}
\usepackage{pifont}
\usepackage{soul, color, xcolor}
\usepackage{colortbl}
\soulregister\cite7
\hypersetup{
colorlinks=true,
linkcolor=black,
citecolor=black
}
\journal{Nuclear Physics B}

\begin{document}

\begin{frontmatter}



\title{HyperTea: \texorpdfstring{Hypergraph-Based}{Hypergraph-Based} Temporal Enhancement and Alignment Network for Moving Infrared Small Target Detection}
%% Authors
\author[label1,label2]{Zhaoyuan Qi}
\ead{qizhaoyuan24@mails.ucas.ac.cn}

\author[label1,label3]{Weihua Gao}
\ead{gaoweihua22@mails.ucas.ac.cn}

\author[label1]{Wenlong Niu\corref{cor1}}
\ead{niuwenlong@nssc.ac.cn}

\author[label1]{Jie Tang}
\ead{tangjie@nssc.ac.cn}

\author[label1]{Yun Li}
\ead{liyun02@nssc.ac.cn}

\author[label1]{Xiaodong Peng\corref{cor1}}
\ead{Pxd@nssc.ac.cn}

\cortext[cor1]{Corresponding author}

%% Affiliations
\affiliation[label1]{organization={Key Laboratory of Electronics and Information Technology for Space Systems, National Space Science Center, Chinese Academy of Sciences},
             city={Beijing},
             postcode={100190},
             country={China}}
\affiliation[label2]{organization={School of Advanced Interdisciplinary Studies, University of Chinese Academy of Sciences},
             city={Beijing},
             postcode={101408},
             country={China}}
\affiliation[label3]{organization={School of Computer Science and Technology, University of Chinese Academy of Sciences},
             city={Beijing},
             postcode={100190},
             country={China}}
% \author{Zhaoyuan Qi, Weihua Gao, Wenlong Niu*, Jie Tang, Yun Li, Xiaodong Peng}
% \affiliation{organization={Key Laboratory of Electronics and Information Technology for Space Systems, National Space Science Center, Chinese Academy of Sciences},
%             city={Beijing},
%             postcode={100190}, 
%             country={China}}

\begin{abstract}
In practical application, moving infrared small target detection (MIRSTD) remains highly challenging owing to the target's small size, low intensity, and complex motion pattern. Existing methods typically model low-order correlations between features and perform feature extraction and enhancement within a single temporal scale. Although hypergraphs have been widely used for high-order correlation learning, they have received limited attention in MIRSTD. To leverage the potential of hypergraphs and enhance multi-timescale feature representation, we propose HyperTea, a framework that integrates global and local temporal perspectives to effectively model high-order spatiotemporal correlations of features. HyperTea consists of three modules: the global temporal enhancement module (GTEM), which realizes global temporal context enhancement through semantic aggregation and propagation; local temporal enhancement module (LTEM) is designed to capture local motion patterns between adjacent frames and then enhance local temporal context; temporal alignment module (TAM) to address potential cross-scale feature misalignment. To the best of our knowledge, HyperTea is the first to integrate convolutional, recurrent, and hypergraph neural networks for MIRSTD, significantly improving detection performance. Experiments on DAUB and IRDST demonstrate its state-of-the-art performance. Code and weights will be available here https://github.com/Lurenjia-LRJ/HyperTea.
\end{abstract}

%% Keywords

\begin{graphicalabstract}
\begin{figure*}
\centering
\includegraphics[width=1\linewidth]{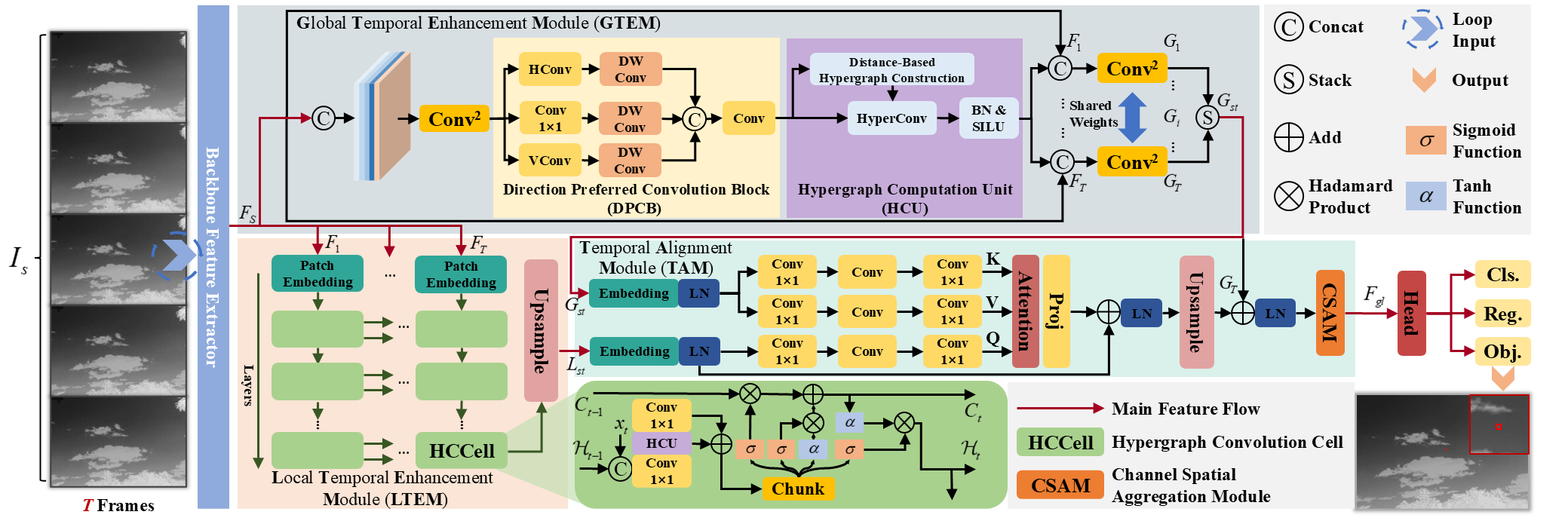}
% \caption{Overview of the proposed framework HyperTea. Our HyperTea consists of a backbone and three key modules: the global temporal enhancement module, local temporal enhancement module, and temporal alignment module. $\mathbf{I}_s$ is $T$ consecutive frames. It is fed into backbone to get spatial feature $\mathbf{F}_s$. GTEM and LTEM enhance $\mathbf{F}_s$ at  global and local temporal scales respectively, yielding $\mathbf{G}_{st}$ and $\mathbf{L}_{st}$. Then $\mathbf{G}_{st}$ and $\mathbf{L}_{st}$ are aligned by TAM to generate $\mathbf{F}_{gl}$. Finally, the fully fused features $\mathbf{F}_{gl}$ are fed into the detection head.} 
\end{figure*}
\end{graphicalabstract}

\begin{highlights}
% \begin{itemize}
    \item Integrates convolutional, recurrent, and hypergraph neural networks into a framework
    \item Proposes dual-scale temporal modules to enhance global and local motion context
    \item Designs a temporal alignment module to mitigate cross-scale feature misalignment
    \item Achieves state-of-the-art on DAUB and IRDST datasets
% \end{itemize}
\end{highlights}

\end{frontmatter}


\section{Introduction}
\label{sec:intro}
Infrared small target detection (IRSTD) has been widely applied in fields such as reconnaissance, surveillance, and rescue~\cite{kouInfraredSmallTarget2023}. In many practical scenarios, unmanned aerial vehicles (UAVs) represent the primary detection targets. However, the inherent characteristics of infrared imaging systems impose the following challenges on IRSTD:
\begin{itemize}
\item \textbf{Dim-small}: The long imaging distance causes a small target scale, weak intensity, low contrast, and low signal-to-clutter ratio (SCR), making targets easily obscured in complex backgrounds.

\item \textbf{Poor spatial distinguishability}: Owing to the inherent characteristics of infrared imaging systems, targets suffer from the absence of color and texture features and exhibit blurred edges. These factors significantly degrade the spatial distinguishability of targets from clutter.

\item \textbf{Spatiotemporal instability}: For moving infrared small targets, driven by factors such as distance changes, pose adjustments, or partial occlusions by clutter, their spatial characteristics (e.g., size and intensity) may exhibit dynamic variations across consecutive frames. Concurrently, their motion features (e.g., velocity and trajectory) are influenced by two primary factors: the intrinsic motion of targets and  imaging platform's motion and perspective transformations. These two factors collectively induce non-uniform and non-linear motion patterns of targets, thereby posing substantial challenges to feature representation learning.

\item \textbf{Target-clutter spatiotemporal coupling}: Background clutter often bears striking similarities to targets in two critical dimensions. Spatially, their features within single frames are similar to those of targets. Temporally, their motion patterns may exhibit short-term similarity to targets', particularly when the imaging platform changes position and orientation, where the relative motion of both targets and clutter with respect to the imaging platform becomes highly analogous. This dual similarity, encompassing spatial characteristics in a single frame and motion trends across consecutive frames, induces strong spatiotemporal coupling between clutter and real targets. Such coupling obscures the discriminative cues between the two, thereby considerably increasing the difficulty of accurate detection.
\end{itemize}

To address these challenges, many IRSTD methods have been proposed, typically categorized into model- and learning-based methods. Model-based methods leverage prior knowledge derived from image observation and analysis to detect targets~\cite{chenLocalContrastMethod2014,zengDesignTopHatMorphological2006,gaoInfraredPatchImageModel2013}. While they perform well in certain scenarios~\cite{zhangInfraredSmallTarget2019,gaoDimSmallTarget2024a}, their heavy reliance on manually designed feature extractors and hyperparameter tuning limits their adaptability.

Contrarily, learning-based methods have recently shown superior detection performance and generalizability~\cite{daiAsymmetricContextualModulation2021,liDenseNestedAttention2023,chenSSTNetSlicedSpatioTemporal2024}.
These methods primarily rely on convolutional neural networks (CNNs)~\cite{heDeepResidualLearning2016a,krizhevskyImageNetClassificationDeep2017}, Transformers~\cite{carionEndtoEndObjectDetection2020,dosovitskiyImageWorth16x162020,liuVideoSwinTransformer2022}, recurrent neural networks (RNNs)~\cite{shiConvolutionalLSTMNetwork2015}, and their hybrids. CNN-based methods~\cite{liDenseNestedAttention2023,wuUIUNetUNetUNet2023a,zhangAttentionGuidedPyramidContext2023b} can effectively extract local features but are constrained by grid-based representations, limiting their ability to model spatial global dependencies. Because transformers overcome this limitation by modeling images as fully connected graphs, some methods use them to improve long-range modeling ~\cite{yuanSCTransNetSpatialChannelCross2024c,tongSTTransSpatialTemporalTransformer2024,linIRTransDetInfraredDim2023a}.
However, the quadratic computational complexity and the loss of low-level details caused by patch embedding may limit their application. RNN-based methods have gained increasing attention owing to their excellent ability in modeling sequential dependencies~\cite{chenSSTNetSlicedSpatioTemporal2024,chenDenseMovingInfrared2024}. Nonetheless, they suffer from difficulties in parallel training and high computational costs, particularly for long sequences. Consequently, despite the diversity of these learning-based architectures, their inherent biases---CNNs' local receptive fields, Transformers' loss of low-level details, and RNNs' high computational costs---restrict their performance in practical scenarios. Additionally, these paradigms are confined to the domain of low-order feature correlation modeling. Such low-order representations are insufficient to capture the complex many-to-many cross-temporal associations among global feature nodes, thereby preventing a single architecture from effectively resolving all aforementioned challenges.

% Recently, Hypergraph Neural Networks (HGNNs) present huge potential in general target detection owing to their higher-order learning ability~\cite{gaoHGNNGeneralHypergraph2023a,fengHyperYOLOWhenVisual2025}, but their application in IRSTD remains to be explored.

% Given that each network architecture has inherent limitations and advantages, and existing methods struggle to model semantic associations across different temporal scales, we naturally conceive the idea of exploring a new paradigm for MIRSTD by leveraging complementary advantages of networks and thus achieve multi-timescale learning. 

CNNs, Transformers, and RNNs are primarily constrained to low-order correlation modeling, while hypergraph neural networks (HGNNs)~\cite{gaoHGNNGeneralHypergraph2023a,fengHyperYOLOWhenVisual2025} have achieved remarkable success in general object detection owing to their high-order learning capabilities. Therefore, we naturally envisioned exploring a new paradigm for MIRSTD by integrating the complementary strengths of the aforementioned networks. This approach enables the learning of high-order feature representations across multiple temporal scales, thereby enhancing overall detection performance.

Therefore, a key problem is how to build a detection pipeline based on fundamental network architectures. Because CNNs excel at efficiently extracting spatial features and preserving low-level details, and RNNs can excellently model sequential dependencies to handle short-term spatiotemporal instability, our framework strategically integrated these two fundamental architectures for global and local temporal context enhancement, respectively. Furthermore, we incorporate HGNNs to endow both global and local features with higher-order dependency representation capabilities, thereby transcending the limitations of conventional low-order modeling.

To fully leverage the complementary advantages of various basic network architectures and promote cross-temporal feature representation learning of targets, we propose a hypergraph-based temporal enhancement and alignment network (HyperTea). The core motivation of HyperTea is to resolve the spatiotemporal instability and target-clutter coupling in MIRSTD---challenges that are often insurmountable for single-architecture models, primarily owing to their lack of higher-order learning capabilities and their limitation to feature representation at a single temporal scale. Specifically, our work focuses on representing infrared targets simultaneously at both global and local temporal scales, particularly using hypergraphs to enable the learning of high-order correlations. The uniqueness of this tri-architecture fusion lies in its synergistic integration, where HGNNs model complex many-to-many associations among feature nodes, while the tailored RNN+HGNN and CNN+HGNN sub-structures enhance motion context at local and global scales, respectively, thereby ensuring a balance between detection performance and computational complexity. To the best of our knowledge, this is the first study to integrate CNNs, RNNs, and HGNNs for MIRSTD. In summary, the main contributions of the study are as follows.

\begin{enumerate}
\item[1)] We explore and propose a pioneering three-architecture scheme, HyperTea, to leverage the multi-timescale feature representation learning of MIRSTD, which first integrates CNNs, RNNs, and HGNNs into a unified framework. 

\item[2)] With HGNNs being applied to IRSTD for the first time, we propose two modules to enhance global and local temporal context. Specifically, based on CNN and HGNN, a global temporal enhancement module (GTEM) was developed to extract high-order semantic correlations from the global temporal context and propagate them to each frame for enhancement. Additionally, we improved ConvLSTM by HGNN and thus designed the local temporal enhancement module (LTEM) to enhance local inter-frame correlations of small targets.

\item[3)] A temporal alignment module (TAM) was designed to diminish the potential misalignment between features across different temporal scales, and amplify the semantic differences between target and background.
\end{enumerate}
%-------------------------------------------------------------------------

%-------------------------------------------------------------------------

\section{Related Work}
\subsection{Hypergraph Learning}
As shown in ~\cite{gaoHypergraphLearningMethods2022a,gaoHGNNGeneralHypergraph2023a}, hypergraphs exhibit impressive capabilities in modeling complex higher-order associations and have been applied in various fields ~\cite{yangLBSN2VecHeterogeneousHypergraph2022,vinasHypergraphFactorizationMultitissue2023,xiaoMultiHypergraphLearningBasedBrain2020}. Recently, general hypergraph neural network ~\cite{gaoHGNNGeneralHypergraph2023a} introduces a spatial approach for higher-order message propagation among feature vertices and achieves remarkable detection performance. Despite these advancements, the application of hypergraph learning remains unexplored in IRSTD. 

\subsection{Infrared Small Target Detection}
According to different driven paradigms, IRSTD methods can be categorized into model- and learning-based methods. 
Model-based methods can be further divided into filter-based methods~\cite{deshpandeMaxmeanMaxmedianFilters1999,zengDesignTopHatMorphological2006}, local contrast-based methods~\cite{chenLocalContrastMethod2014,hanInfraredSmallTarget2018,hanInfraredSmallTarget2021,hanLocalContrastMethod2020}, and low-rank representation-based methods~\cite{gaoInfraredPatchImageModel2013, zhangInfraredSmallTarget2019}. These methods typically rely on specific hypotheses, manually designed feature extractors, and hyperparameters based on prior knowledge. Such dependencies often limit their generalizability in complex real-world scenarios~\cite{chenSSTNetSlicedSpatioTemporal2024}. However, learning-based methods achieve superior detection performance owing to their powerful feature representation and learning ability. Depending on the number of reference frames used, learning-based methods can be further classified into single-frame and multi-frame schemes.

As a prominent research direction, single-frame schemes have recently attracted considerable attention. CNN-based methods are typically built upon U-Net~\cite{ronnebergerUNetConvolutionalNetworks2015} and its variants, with feature fusion serving as a representative improvement strategy. For example, ACM~\cite{daiAsymmetricContextualModulation2021} realizes feature fusion of different levels through asymmetric context modulation. DNANet~\cite{liDenseNestedAttention2023} achieves better detection performance through gradual interaction and enhancement between high- and low-level features. To reach multilevel and multiscale representation learning, UIUNet~\cite{wuUIUNetUNetUNet2023a} embeds a tiny U-Net into a larger U-Net backbone. ISNet~\cite{zhangISNetShapeMatters2022} extracts the underlying edge features and improves detection performance through cross-layer feature fusion. RDIAN~\cite{sunReceptiveFieldDirectionInduced2023} utilizes multiscale convolution and direction-induced attention mechanisms to capture richer target features for fusion. PGNet~\cite{liPGNetPositionGuided2025} employs a cascaded convolutional encoder architecture for the adaptive extraction of multiscale discriminative infrared features. Recently, many studies have combined CNNs with ViTs~\cite{dosovitskiyImageWorth16x162020} to leverage their substantial advantage in modeling long-range dependencies. IR-TransDet~\cite{linIRTransDetInfraredDim2023a} explores the relationship between the global image, target, and neighboring pixels. SCTransNet~\cite{yuanSCTransNetSpatialChannelCross2024c} efficiently encodes semantic differences between target and background through spatial channel cross-transformers. Inspired by the neural ordinary differential equations, RKformer~\cite{zhangRKformerRungeKuttaTransformer2022} and ABMNet~\cite{chenABMNetCouplingTransformer2023} were proposed. Considering the high labeling costs of full supervision, 
point-supervision schemes~\cite{liMonteCarloLinear2023b,yingMappingDegenerationMeets2023,yangLabelEvolutionBased2024a,liLevelSetAnnotation2024,kouMCGCMultiscaleChain2024,wuSinglePointSupervisedHighResolution2024} and self-supervision schemes~\cite{luSIRST5KExploringMassive2024} have emerged as viable alternatives.
Recently, some popular network architectures such as Mamba~\cite{chenMiMISTDMambainMambaEfficient2024a} and SAM~\cite{zhangIRSAMAdvancingSegment2025} have also migrated to IRSTD. Moreover, FEST~\cite{zhaoRobustInfraredSmall2025a} presents a feature-enhanced and sensitivity-tunable framework to improve the detection performance of existing single-frame methods. Meanwhile, SGIRNet~\cite{chiContrastinvariantFeatureExtraction2025} focuses on improving the generalizability of existing models through framework-level improvements.
However, the lack of motion information limits single-frame schemes in detecting moving infrared targets under complex dynamic backgrounds. To overcome this problem, some multi-frame strategies have been proposed recently.

Depending on the primary network architecture, multi-frame schemes can be further categorized into CNN-based, CNN with Transformer, and CNN with RNN approaches. CNN-based schemes such as STDMANet~\cite{yanSTDMANetSpatioTemporalDifferential2023b} propose spatiotemporal differential multiscale attention networks. DTUM~\cite{liDirectionCodedTemporalUShape2023a} uses direction-coded convolution to extract target motion features. TMP~\cite{zhuTMPTemporalMotion2024c} extracts temporal and spatial features in parallel and performs cross-domain feature fusion. As for CNN with Transformer methods, ST-Trans~\cite{tongSTTransSpatialTemporalTransformer2024} models spatiotemporal correlations through an improved Video Swin Transformer~\cite{liuVideoSwinTransformer2022}. Tridos~\cite{duanTripleDomainFeatureLearning2024a} uses Video Swin Transformer~\cite{liuVideoSwinTransformer2022} to capture global frequency-domain features. To address the issue of insufficient feature interaction, TSINet~\cite{zhuangTemporalSemanticInteractionNetwork2025b} proposes a multi-scale semantic interaction Transformer to enable robust cross-level feature fusion. To exploit the powerful temporal modeling ability of RNNs, CNN with RNN schemes such as SSTNet~\cite{chenSSTNetSlicedSpatioTemporal2024} models spatiotemporal features in cross-slice motion context based on improved ConvLSTM~\cite{shiConvolutionalLSTMNetwork2015}. Furthermore, in terms of weakly supervised learning, S2MVP~\cite{duanSemiSupervisedMultiviewPrototype2025a} achieves remarkable performance using a small number of labeled training samples. To utilize prior knowledge in linguistic cues, visual-language detection frameworks such as MoPKL~\cite{chenMotionPriorKnowledge2025} are gradually attracting attention. 

However, current IRSTD methods have not yet involved cross-temporal scales (local-global time range) context modeling. Additionally, existing methods can only capture low-order correlations between feature nodes and fail to model correlations among multiple nodes. Modeling high-order correlations between nodes and utilizing them to enhance motion context remains unexplored. 

Hyperspectral target tracking methods~\cite{zhuDSPNetDynamicSpectral2023a,zhaoSASUNetHyperspectralVideo2025a,zhaoSiamBSIHyperspectralVideo2025,zhaoSiamSTUHyperspectralVideo2025} recently explore mechanisms such as template updating and scale-aware adaptation to facilitate robust representation learning for moving targets. While these advancements provide valuable insights for MIRSTD, there is a fundamental discrepancy in their task settings: tracking-based methods rely on explicit target initialization, whereas MIRSTD demands direct detection without any prior target information. Consequently, these methods serve as contextual motivation for our feature enhancement design rather than as direct baselines for performance comparison.

Therefore, we focus on target feature enhancement and alignment across temporal scales, particularly using HGNNs to learn high-order correlations. For intuitiveness and clarity, we summarize the unique advantages of HyperTea:

\begin{enumerate}
\item[1)] To overcome the limitation of learning-based multi-frame methods that capture infrared target features at a single temporal scale, we consider enhancing target features at both local and global temporal scales. 

\item[2)] Conventional networks, such as CNNs and Transformers can only model low-order relationships between nodes. Considering the complex many-to-many relationships among nodes, we introduce HGNNs into IRSTD.

\item[3)] Conventional learning-based methods fail to leverage the complementary advantages among basic network structures. Given this, we design a new network, HyperTea,  to address it.
\end{enumerate}

\section{Methodology}
\subsection{Preliminaries}
Compared to ordinary graphs (such as Transformers) that are limited to modeling pair-wise dependencies, the most important feature of hypergraphs is that hyperedges can be associated with multiple vertices, which makes it possible to model more complex higher-order relationships.
A hypergraph~\cite{gaoHGNNGeneralHypergraph2023a} can be defined as $\mathcal{G=\{V,E\}}$. 
Given a flattened feature map $\mathbf{X}$, we treat each feature point as a vertex, the collective of which constitutes the vertex set $\mathcal{V}$. 
To construct hyperedge set $\mathcal{E}$, any group of vertices whose pairwise semantic distances fall below a predefined threshold $\epsilon$ is grouped into a hyperedge $e$. 
The process is mathematically formulated as: $\mathcal{E} = \{e(v, \epsilon) \mid v \in \mathcal{V}\}$, where $e(v, \epsilon) = \{u \mid \| \boldsymbol{x}_u - \boldsymbol{x}_v \|_2 < \epsilon, u \in \mathcal{V}\}$ indicate the hyperedge centered at the specified vertex $v$. $\| \boldsymbol{x_i} - \boldsymbol{x_j} \|_2$ is the
 Euclidean distance function.

Based on  the spatial-domain hypergraph convolution with extra residual connections~\cite{fengHyperYOLOWhenVisual2025}, a hypergraph convolution unit (HCU) can be formulated as:
\begin{align}
\text{HCU}(\mathbf{X}) &= \text{HyperConv}(\mathbf{X},\mathbf{H})\notag{}\\
&= \mathbf{X} + \mathbf{D}_v^{-1} \mathbf{H} \mathbf{D}_e^{-1} \mathbf{H}^T \mathbf{X} \boldsymbol{\Theta}
\tag{1}
\end{align}
where H represents the incidence matrix of X, $D_v$ and $D_e$ represent the diagonal degree matrices of the vertices and hyperedges respectively, $\Theta$ is a fully connected layer. 

Hypergraph convolution can be understood as follows. The multiplication of $\mathbf{H}^T$ facilitates the information aggregation from nodes to hyperedges, followed by normalization using $\mathbf{D}_e$. Subsequently, multiplying with $\mathbf{H}$ aggregates information from hyperedges back to vertices, with normalization performed via $\mathbf{D}_v$. Through this two-stage information propagation, hypergraph convolution can model the high-order correlations between vertices.

\subsection{Overall Architecture}
\begin{figure*}
\centering
\includegraphics[width=1\linewidth]{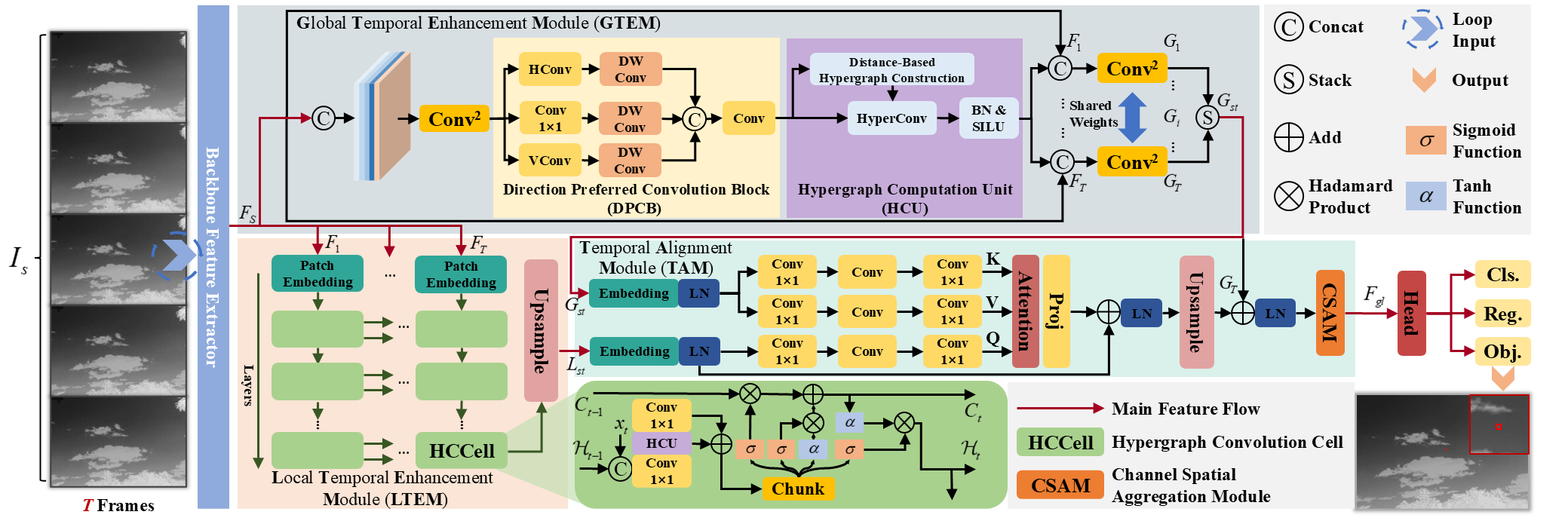}
\caption{Overview of the proposed framework HyperTea. Our HyperTea consists of a backbone and three key modules: the global temporal enhancement module, local temporal enhancement module, and temporal alignment module. $\mathbf{I}_s$ is $T$ consecutive frames. It is fed into backbone to get spatial feature $\mathbf{F}_s$. GTEM and LTEM enhance $\mathbf{F}_s$ at  global and local temporal scales respectively, yielding $\mathbf{G}_{st}$ and $\mathbf{L}_{st}$. Then $\mathbf{G}_{st}$ and $\mathbf{L}_{st}$ are aligned by TAM to generate $\mathbf{F}_{gl}$. Finally, the fully fused features $\mathbf{F}_{gl}$ are fed into the detection head.} 
\label{fig:arch}
\end{figure*}
\begin{figure}
\centering
\includegraphics[width=1\linewidth]{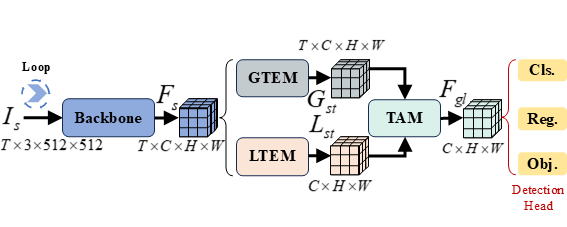}
\caption{Simplified workflow of our HyperTea. It contains the calculation process of input $\mathbf{I}_s$ and the feature flow of target detection.} 
\label{fig:wf}

\end{figure}
We aim to effectively improve the network's ability to detect moving small targets through cross-temporal feature learning. To achieve this goal, we propose a new framework, as shown in~\Cref{fig:arch}. For clarity, its simplified workflow is presented in~\Cref{fig:wf}.

Our HyperTea starts with $T$ consecutive frames $\mathbf{I}_s = [\mathbf{I}_1, \mathbf{I}_2, \ldots, \mathbf{I}_T] \in \mathbb{R}^{T \times 3 \times 512 \times 512}$ as input and outputs the detection results on keyframe \textit{$\mathbf{I}_T$}. 
Following the common paradigm of MIRSTD~\cite{chenSSTNetSlicedSpatioTemporal2024,zhuTMPTemporalMotion2024c,duanTripleDomainFeatureLearning2024a}, we use the CSPDarknet~\cite{geYOLOXExceedingYOLO2021a} as the backbone. To realize the strategy of cross-temporal feature learning, we design three key modules: GTEM, LTEM, and TAM.

As for the pipeline, we first feed each frame into the backbone with shared weights to obtain multi-frame spatial features $\mathbf{F}_{s} = \{\mathbf{F}_1, \mathbf{F}_2, \ldots, \mathbf{F}_T\} \in \mathbb{R}^{T \times C \times H \times W}$, where $C$, $H$, and $W$ denote the channel, height, and width of the feature map, respectively. In our implementation, we set $C=128$, $H=64$, and $W=64$. Subsequently, GTEM receives $\mathbf{F}_{s} \in \mathbb{R}^{T \times C \times H \times W}$ as its input and outputs the global temporal context-enhanced feature $\mathbf{G}_{st} = \{\mathbf{G}_1, \mathbf{G}_2, \ldots, \mathbf{G}_T\} \in \mathbb{R}^{T \times C \times H \times W}$. LTEM also receives $\mathbf{F}_{s} \in \mathbb{R}^{T \times C \times H \times W}$ as its input and outputs the local temporal context-enhanced feature $\mathbf{L}_{st} \in \mathbb{R}^{C \times H \times W}$. To achieve the aggregation of the global and local temporal context , we then feed $\mathbf{G}_{st} \in \mathbb{R}^{T \times C \times H \times W}$ and $\mathbf{L}_{st} \in \mathbb{R}^{C \times H \times W}$ into TAM to mitigate misalignment at different temporal scales and get the fully fused features $\mathbf{F}_{gl} \in \mathbb{R}^{C \times H \times W}$. The above pipeline can be formulated as follows:
\begin{align*}
\begin{cases}
\mathbf{G}_{st} &= \operatorname{GTEM}(\mathbf{F}_{s}) \\
\mathbf{L}_{st} &= \operatorname{LTEM}(\mathbf{F}_{s}) \\
\mathbf{F}_{gl} &= \operatorname{TAM}(\mathbf{G}_{st}, \mathbf{L}_{st}) \tag{2}
\end{cases}
\end{align*}
Finally, the global---local aligned temporal feature $\mathbf{\textit{F}}_{gl}$ is fed into the detection head to obtain the detection results.

\subsection{Global Temporal Enhancement Module}
Enhancing global temporal context through inter-frame dependencies remains a challenging problem in MIRSTD. 
Previous approaches usually use the attention mechanism to model spatial dependencies~\cite{duanTripleDomainFeatureLearning2024a} or inter-channel semantic dependencies~\cite{zhuTMPTemporalMotion2024c}. 
However, because the attention mechanism essentially models the image as a fully connected graph, the limited representation capability hinders higher-order dependency learning. 
Recently, global---local semantic interactions have emerged as an effective paradigm for feature enhancement~\cite{yuanSCTransNetSpatialChannelCross2024c,fengHyperYOLOWhenVisual2025}.
Inspired by these, we propose GTEM as shown in~\Cref{fig:arch}, which includes temporal feature aggregation, direction-preferred convolution, higher-order association learning, and global semantic scattering mechanism. 

Specifically, we first aggregate global temporal feature through two convolution layers.  
\begin{equation}
\mathbf{L}_{gt} = \text{Conv}_{3\times3}^2(\text{Concat}[\mathbf{F}_{1},...,\mathbf{F}_{T}])) \tag{3}
\end{equation}
where $\text{Conv}_{3\times3}^2$ means two ${3\times3}$ basic convolution components with batch normalization and an activation layer.

Practically, features of background clutter often exhibit striped or bar-like patterns corresponding to physical objects. Therefore, we design a direction-preferred convolution block (DPCB) to encode such discriminative features into the channel dimension to enhance the feature representation, as shown in~\Cref{fig:arch}. In terms of implementation, we use convolution with kernel sizes $1\times5$, $5\times1$, and $1\times1$ to extract features with different direction preferences. Subsequently $3\times3$ basic depth-wise convolution components are used to enhance the local spatial features and expand the receptive field. Thereafter, we apply $3\times3$ convolution to fuse the extracted features and obtain $D_{gt}$.
\begin{align*}
\begin{cases}
\mathbf{L}_{h} = \text{DWC}(\text{Conv}_{1\times5}(\mathbf{L}_{gt})) \\
\mathbf{L}_{v} = \text{DWC}(\text{Conv}_{5\times1}(\mathbf{L}_{gt})) \\
\mathbf{L}_{s} = \text{DWC}(\text{Conv}_{1\times1}(\mathbf{L}_{gt})) \\
\mathbf{D}_{gt} = \text{Conv}_{3\times3}(\text{Concat}[\mathbf{L}_{h},\mathbf{L}_{v},\mathbf{L}_{s}]) +\mathbf{L}_{gt}
\tag{4}
\end{cases}
\end{align*}
where DWC implies $3\times3$ depth-wise convolution, $\mathbf{L}_{h},\mathbf{L}_{v},\mathbf{L}_{s}$ represent features with horizontal direction preference, vertical direction preference, and no directional preference, respectively.

Subsequently, HCU receives $D_{gt}$ as its input and generates higher-order semantic-enhanced feature $G_{h}$. To enhance global temporal context through $G_{h}$, we propagate $G_{h}$ to each time step through convolution layers with shared weights. 
Therefore, the global temporal enhanced features $G_{st}$ can be obtained as follows:
\begin{align*}
\mathbf{G}_{h} &= \text{HCU}(\mathbf{D}_{gt}) 
\tag{5}\\
\mathbf{G}_{i} & = \text{Conv}_{3\times3}^2(\text{Concat}[\mathbf{F}_{i},\mathbf{G}_{h}]), i=1,...,T
\tag{6}
\end{align*}

\subsection{Local Temporal Enhancement Module}
To achieve robust and accurate detection on keyframe, we  that it is crucial to effectively model local motion patterns, particularly when addressing dynamic complex scenes. Considering the powerful capability of convLSTM~\cite{shiConvolutionalLSTMNetwork2015} and its variants~\cite{wangPredRNNRecurrentNeural2023,tangSwinLSTMImprovingSpatiotemporal2023,chenSSTNetSlicedSpatioTemporal2024,chenDenseMovingInfrared2024} in spatiotemporal representation learning, in combination with HGNNs' advantages in higher-order learning, we designed a hypergraph convolution LSTM and its basic cell (HCCell). 
ConvLSTMs are inherently constrained by local receptive fields, limited to modeling low-order, grid-based spatial correlations. Contrarily, the HCU within the HCCell is designed to capture global high-order semantic dependencies. This transition from local, low-order feature extraction to global, high-order relational learning allows HCCell to enhance its local spatiotemporal feature learning capability.

We made the following improvements based on the ConvLSTM node. As shown in~\Cref{fig:arch}, we first project the input $\mathcal{X}_t$ and hidden state $\mathcal{H}_{t-1}$ into the same semantic space using ${1\times1}$ convolution. Then, we perform high-order semantic enhancement by HCU to obtain the enhanced feature $\mathcal{L}_{h}$. Subsequently, we apply convolution with residual connections to achieve semantic modulation of $\mathcal{L}_{h}$ and obtain $\mathcal{F}_t$. Subsequently, we split $\mathcal{F}_t$ along the channel dimension and then apply corresponding activation functions to finally obtain four gates. The computational model of the HCCell is presented mathematically as follows:
\begin{align}
\begin{cases}
\mathcal{L}_h =\text{HCU}(\text{Conv}_{1\times1}([\mathcal{X}_t,\mathcal{H}_{t-1}])) \\
\mathcal{F}_t = \text{Conv}_{1\times1}(\mathcal{L}_h)+\mathcal{L}_{h} \\
[{i}_t, {f}_t, {o}_t, {g}_t] 
= [\sigma, \sigma, \sigma, \tanh](\text{Chunk}(\mathcal{F}_t))\\
\mathcal{C}_t = {f}_t \circ \mathcal{C}_{t-1} + {i}_t \circ {g}_{t} \\
\mathcal{H}_t = {o}_t \circ \tanh(\mathcal{C}_t) 
\tag{7}
\end{cases}
\end{align}
where $[\sigma, \sigma, \sigma, \tanh]$ denote the activation functions for ${i}_{t}, {f}_t, {o}_t$ and ${g}_t$ respectively, and $\circ$ denotes the Hadamard (element) product.

\subsection{Temporal Alignment Module}
Our approach utilizes different branches to leverage the global and local spatiotemporal features representations of infrared targets. However, the relative motion between target and imaging platform along with the learning path differences often leads to feature misalignment between global and local spatiotemporal features. To address this, we propose TAM as illustrated in~\Cref{fig:arch}, which consists of global---local temporal attention (GLTA) and channel spatial aggregation module (CSAM).

1) Global-Local Temporal AttentionJ(GLTA): Specifically, we first perform patch embedding and layer normalization on global temporal 
context-enhanced feature $\mathbf{G}_{st}$, and  local temporal 
context-enhanced feature $\mathbf{L}_{st}$. 
\begin{align}
% \begin{equation}
\hat{\mathbf{G}}_{st} &=\text{LN}(\text{EMD}(\mathbf{G}_{st})) \tag{8} \\
\hat{\mathbf{L}}_{st} &=\text{LN}(\text{EMD}(\mathbf{L}_{st})) \tag{9}
\end{align}

Subsequently, we use $1\times1$ convolution to project $\mathbf{G}_{st}$ and $\mathbf{L}_{st}$ into semantic spaces with the same dimensions. Then, in order to fuse features at different temporal scales, our GLTA employs $\mathbf{L}_{st}$ as query, $\mathbf{G}_{st}$ as key and value to realize fine-grained feature fusion.
\begin{align}
\mathbf{Q} &= \text{Conv}_{1\times1}(\text{Conv}_{3\times3}(\text{Conv}_{1\times1}(\hat{\mathbf{L}}_{st}))) \tag{10}\\
\mathbf{K} &= \text{Conv}_{1\times1}(\text{Conv}_{3\times3}(\text{Conv}_{1\times1}(\hat{\mathbf{G}}_{st}))) \tag{11}\\
\mathbf{V} &= \text{Conv}_{1\times1}(\text{Conv}_{3\times3}(\text{Conv}_{1\times1}(\hat{\mathbf{G}}_{st}))) \tag{12}\\
Attn &= 
\text{Softmax}(\frac{\mathbf{Q}\mathbf{K}^T}{\sqrt{d}})\mathbf{V} \tag{13}
\end{align}
where $\sqrt{d}$ is a normalized scale factor.
In this way, any two pixels can interact, and cross-temporal dependencies between them can be captured, rather than being confined solely to the current temporal scale.

In addition, to preserve more original information, we also incorporate residual blocks to embed keyframe features at different temporal scales into the query results $\mathbf{R}$:
\begin{align}
\mathbf{R} &= \text{LN}(\mathbf{G}_{T}+\text{UP}(\text{LN}(\text{Conv}_{1\times1}(Attn)+\hat{\mathbf{L}}_{st}))) \tag{14}
\end{align}
where LN is the layer normalization, UP denotes the reconstruction layer consisting of an upsample layer and a convolution layer.

2) Channel Spatial Aggregation Module(CSAM): Recently, leveraging channel and spatial global information has become a common practice in feature enhancement. Inspired by~\cite{liDenseNestedAttention2023,yuanSCTransNetSpatialChannelCross2024c}, 
we propose CSAM to further optimize the selection of spatial and channel weights.
\begin{figure}
    \centering
    \includegraphics[width=0.5\linewidth]{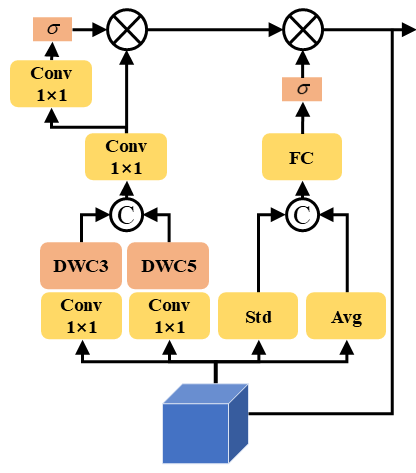}
    \caption{Details of our proposed CSAM.}
    \label{fig:CSAM}
    
\end{figure}

As shown in~\Cref{fig:CSAM}, given an input tensor $\mathbf{X} \in \mathbb{R}^{C \times H \times W}$, CSAM first adjusts the channel dimension by $1\times 1$ convolution and employs $3\times3$ and $5\times5$ depth-wise convolution to enhance the local spatial information. CSAM then compresses the channel dimension to 1 to obtain the spatial attention map $\mathbf{X}_{sa}$. The above process can be formulated as:
\begin{align}
\mathbf{X}_{s} &= f_{1}^c([f_{3}^{dwc}(f_{1}^c(\mathbf{X})),f_{5}^{dwc}(f_{1}^c(\mathbf{X}))])\tag{15}\\
\mathbf{X}_{sa} &= f_{sa}^c(\mathbf{X}_{s}) \odot \mathbf{X}_{s}\tag{16}
\end{align}
where $f_{1}^c$ denotes $1\times1$ convolution, $f_{3}^{dwc}$ and $f_{5}^{dwc}$ represent $3\times3$ and $5\times5$ depth-wise convolutions, $f_{sa}^c$ denotes $1\times1$ convolution whose output channel is 1 with sigmoid activation function, and $\odot$ is the broadcasted Hadamard product.

To better estimate the weights of different channels, we compute the maximum value and standard deviation of each channel and further use learnable parameters to obtain channel weights $\mathbf{X}_{ca}$ as follows:
\begin{align}
\mathbf{X}_{ca} &= fc([\text{max}(\mathbf{X}),\text{std}(\mathbf{X})]) \tag{17}
\end{align}
where $fc$ denotes a linear transform layer with sigmoid activation function. 
Finally, CSAM enriches the representation of features by integrating complementary spatial and channel information with residual concatenation.
\begin{align}
\mathbf{X}_{sc} &= \mathbf{X}_{ca} \odot \mathbf{X}_{sa} + \mathbf{X} \tag{18}
\end{align}

\subsection{Loss Function}
Following the general detection paradigm~\cite{geYOLOXExceedingYOLO2021a,chenSSTNetSlicedSpatioTemporal2024}, a conventional loss function can be defined as follows: 
\begin{equation}
\mathcal{L} = \lambda_{\text{reg}} \mathcal{L}_{\text{reg}} + \lambda_{\text{cls}} \mathcal{L}_{\text{cls}} + \lambda_{\text{obj}} \mathcal{L}_{\text{obj}}
\tag{19}
\end{equation}
where $\mathcal{L}_{\text{reg}}$ is the bounding box regression loss, $\mathcal{L}_{\text{cls}}$ is the classification loss, and $\mathcal{L}_{\text{obj}}$ is the target probability loss. $\lambda_{\text{reg}}$, $\lambda_{\text{cls}}$, $\lambda_{\text{obj}}$ are the hyperparameters that balance the loss terms. For $\mathcal{L}_{\text{cls}}$ and $\mathcal{L}_{\text{obj}}$, we use binary cross-entropy loss function. Hyperparameters and $\mathcal{L}_{\text{reg}}$ follow~\cite{duanTripleDomainFeatureLearning2024a}.

\begin{table*}[h!]
\centering
\resizebox{\linewidth}{!}{
\begin{tabular}{c|c|c|c|c|c|c|c|c}
\hline
 \textbf{Dataset} & \textbf{Sequential} & \textbf{Target size} & \textbf{Frames} & \textbf{Image size} & \textbf{Frame rate} & \textbf{Sample} & \textbf{Eff. rate} & \textbf{Scene description} \\ \hline
 & & & & $720 \times 480$ & 30 fps & \checkmark & 10 fps & \\ \cline{5-8}
 \multirow{-2}{*}{IRDST} & \multirow{-2}{*}{\checkmark} & \multirow{-2}{*}{$1 \sim 100$ pixels} & \multirow{-2}{*}{40,650} & $934 \times 696$ & 10 fps & \checkmark & 3.3 fps & \multirow{-2}{*}{\begin{tabular}[c]{@{}c@{}}ground, forest, air, lakes, suburb, clouds, buildings, mountains\end{tabular}} \\ \hline
 DAUB & \checkmark & $1 \sim 16$ pixels & 13,778 & $256 \times 256$ & 100 fps & \ding{55} & 100 fps & ground, forest, air, plain, suburb, air ground junction background\\ \hline
\end{tabular}
}
\captionsetup{justification=centering}
\caption{Details of the IRDST and DAUB datasets. Note that the sampling process is performed by the dataset providers of IRDST to decrease the similarity of images. Eff. rate means the effective frame rate after sampling.}
\label{table:data}
\end{table*}

\begin{table*}[h!]
\centering
\captionsetup{justification=centering} % 设置标题居中

\resizebox{\linewidth}{!}{ 
\begin{tabular}{c|c|c|c|cccc|cccc}
\hline
% --- 第一行：前四列留空，处理右侧合并列 ---
& & & 
& \multicolumn{4}{c|}{\textbf{IRDST}} 
& \multicolumn{4}{c}{\textbf{DAUB}} \\ 
\cline{5-12}

% --- 第二行：在此处定义前四列的 \multirow，并补全右侧指标 ---
\multirow{-2}{*}{\textbf{Scheme}} & 
\multirow{-2}{*}{\textbf{Methods}} & 
\multirow{-2}{*}{\textbf{Publication}} & 
\multirow{-2}{*}{\textbf{Architecture}} & 
\textbf{mAP$_{50}$} & \textbf{Pr} & \textbf{Re} & \textbf{F1} & 
\textbf{mAP$_{50}$} & \textbf{Pr} & \textbf{Re} & \textbf{F1} \\ \hline
\multirow{11}{*}{single-frame}
&ACM~\cite{daiAsymmetricContextualModulation2021}&WACV 2021 & CNN
&65.31 & 82.11 & 80.70& 81.40
& 83.41 & 93.11 &90.65&91.86
\\ 
&ALC~\cite{daiAttentionalLocalContrast2021} &TGRS 2021& CNN
& 53.10 & 77.30 & 69.03 & 72.93
& 79.03&90.41&88.43&89.40
\\ 
&ISNet~\cite{zhangISNetShapeMatters2022}&CVPR 2022 & CNN
&66.30&84.24&79.80&81.96
& 83.02&93.52&89.99&91.72
\\ 
&AGPCNet~\cite{zhangAttentionGuidedPyramidContext2023b}&TAES 2023& CNN
& 67.10 & 85.40 & 79.74& 82.47
& 82.43 & 90.17 &92.62&91.38
\\ 
&DNANet~\cite{liDenseNestedAttention2023} &TIP 2023&CNN
& 71.39 & 88.36 & 81.63 & 84.86
& 84.44 &97.37&87.34&92.08
\\ 
&RDIANet~\cite{sunReceptiveFieldDirectionInduced2023} &TGRS 2023& CNN
& 68.70 & 85.29 &81.86& 83.54
& 80.72 &88.29&92.95&90.56
\\  
&MSHNet~\cite{liuInfraredSmallTarget2024}&CVPR 2024& CNN
&71.04 & 87.92&81.89&84.80
& 85.25 & \textbf{99.30} &86.67&92.56
\\ 
&RPCANet~\cite{wuRPCANetDeepUnfolding} & WACV 2024 &CNN
&68.80& 84.41 & 82.29& 83.33
&78.90&82.85&96.93&89.34
\\ 
&MTUNet~\cite{wuMTUNetMultilevelTransUNet2023} &TGRS 2023& CNN+Trans
&67.20 & 86.35&78.70&82.35
&77.21&93.40&83.42&88.13
\\ 
&SCTransNet~\cite{yuanSCTransNetSpatialChannelCross2024c} &TGRS 2024& CNN+Trans
& 68.41 & 82.93 &83.08&83.01
& 79.97 & 94.34 & 85.99& 89.97
\\ 
&SeRankDet~\cite{daiPickBunchDetecting2024a} &TGRS 2024&CNN+Trans
& 69.50 & 86.16 & 81.38& 83.70
& 84.67 & 94.60 &90.47&92.49
\\ 
\hline
\multirow{7}{*}{multi-frame}
&DTUM~\cite{liDirectionCodedTemporalUShape2023a} &TNNLS 2023&CNN
& 71.80 & 85.12 & \textbf{85.36}& 85.24
& 90.46 & 97.86 &93.58&95.67
\\ 
&STME~\cite{pengMovingInfraredDim2025a} &EAAI 2025&CNN
&69.95&87.64&80.63&83.99
&87.18&95.17&92.85&93.99
\\ 
&SSTNet~\cite{chenSSTNetSlicedSpatioTemporal2024} &TGRS 2024&CNN+RNN
& 70.91  & 87.73 & 81.98 & 84.76
&89.07&97.12&92.72&94.87
\\ 
&TMP~\cite{zhuTMPTemporalMotion2024c} &ESWA 2024&CNN+Trans
& 69.12 & 86.57 & 80.99 & 83.68
& 85.37 & 94.23 & 91.74 & 92.96
\\ 
&Tridos~\cite{duanTripleDomainFeatureLearning2024a} &TGRS 2024&CNN+Trans
& 72.54 & 87.82 & 83.63 & 85.63 
&91.52&96.83&95.52&96.17 
\\ 
&TSINet~\cite{zhuangTemporalSemanticInteractionNetwork2025b} &KBS 2025&CNN+Trans
&73.19&87.45&84.97&86.19 
&88.32&98.46&90.55&94.34
\\ 
&HyperTea(Ours) &-&CNN+RNN+HGNN
&\textbf{76.42}&\textbf{91.18}&84.33&\textbf{87.62} 
% & \textbf{96.03} & 97.79 & \textbf{99.14} & \textbf{98.46} 
& \textbf{95.59} & 98.14 & \textbf{98.31} &\textbf{98.23}
\\  \hline
\end{tabular}
}
\caption{Comparisons with state-of-the-art methods on four metrics: mAP$_{50}$, Precision, Recall, and F1.}
\label{table:cmp}
\end{table*}

\section{Experiments}
\subsection{Datasets and Evaluation Metrics}
To fully validate the effectiveness and superiority of our HyperTea, we conduct experiments on two datasets, DAUB~\cite{720626420933459968} and IRDST~\cite{sunReceptiveFieldDirectionInduced2023}. 
~\mbox{~\Cref{table:data}} shows the profiles of both DAUB and IRDST.
The division of datasets follows~\cite{chenSSTNetSlicedSpatioTemporal2024}.

In terms of evaluation metrics, we follow the widely adopted practice in the object detection paradigm~\cite{geYOLOXExceedingYOLO2021a} and adopt precision (Pr), recall (Re), F1-score, and mean average precision at an IoU threshold of 0.5 (mAP$_{50}$). These metrics are defined as follows:
\begin{align*}
\text{Precision} &= \frac{\text{TP}}{\text{TP} + \text{FP}}, \tag{20} \\
\text{Recall} &= \frac{\text{TP}}{\text{TP} + \text{FN}}, \tag{21} \\
\text{F1} &= \frac{2 \times \text{Precision} \times \text{Recall}}{\text{Precision} + \text{Recall}}, \tag{22}
\end{align*}
where TP, FP, and FN denote the number of true positives (correctly detected targets), false positives (background regions incorrectly identified as targets), and false negatives (missed targets), respectively. The F1-score serves as a harmonic mean of precision and recall, providing a balanced measure that jointly evaluates a detector’s ability to minimize both missed detections and false alarms.
\subsection{Implementation Details}
The input frame resolution is reshaped to 512 × 512 in all experiments. All methods are trained for 50 epochs with a batch size of 4. 
We set the initial learning rate to 0.01, and adopt SGD as the optimizer with momentum 0.937, weight decay $5\times10^{-4}$, and learning rate reduction factor 0.1. 
For non-maximum suppression, the IoU threshold is set to 0.65, whereas the confidence threshold is 0.001. We performed all experiments on one Nvidia GeForce 4090D GPU. 

\subsection{Comparisons With Other Methods} 

1) \textit{Quantitative Comparison:}
Considering that learning-based methods tend to have much superior detection performance compared to model-based ones, we select some state-of-the-art learning-based IRSTD methods for comparison. 
To ensure a comprehensive comparison, we specifically select representative methods from different mainstream architectures, including CNN-based (e.g.,DNANet~\cite{liDenseNestedAttention2023} and DTUM~\cite{liDirectionCodedTemporalUShape2023a}), CNN+Transformer (e.g., SCTransNet~\cite{yuanSCTransNetSpatialChannelCross2024c} and Tridos~\cite{duanTripleDomainFeatureLearning2024a}), and CNN+RNN (e.g., SSTNet~\cite{chenSSTNetSlicedSpatioTemporal2024}). Specifically, to provide a clearer taxonomic perspective on the comparative study, we explicitly present the primary architectural paradigm of each method in\mbox{~\Cref{table:cmp}}. For segmentation-based methods, we follow the paradigm of combined detector~\cite{chenSSTNetSlicedSpatioTemporal2024,duanTripleDomainFeatureLearning2024a}.
In addition, hyperspectral tracking methods such as SiamSTU~\cite{zhaoSiamSTUHyperspectralVideo2025} and SiamBSI~\cite{zhaoSiamBSIHyperspectralVideo2025} focus on target tracking rather than detection. Although hyperspectral tracking methods demonstrate strong capability in modeling temporal consistency, they are developed for tracking tasks that rely on predefined target initialization. In contrast, our HyperTea addresses detection tasks without prior target information, leading to fundamentally different problem settings. Therefore, we do not perform direct comparisons with hyperspectral tracking methods.

The quantitative performance of different detection methods on two datasets is clearly shown in~\Cref{table:cmp}. 
Based on the data in~\Cref{table:cmp}, three important findings can be identified.
%findings 1
First, the detection performance of single-frame methods is typically inferior to those of multi-frame methods; however, they may outperform certain multi-frame approaches when temporal associations within the dataset are challenging to exploit. For instance, on DAUB, the SOTA single-frame method, MSHNet achieves an mAP$_{50}$ of 85.25\% and an F1 score of 92.56\%, which are lower than any multi-frame method listed in~\Cref{table:cmp}. That's because single-frame methods rely solely on spatial features, thereby limiting their capacity to adaptively capture the temporal dynamics of targets. However, on IRDST, the leading single-frame method, DNANet, attains an mAP$_{50}$ of 71.39\% and an F1 score of 84.86\%, surpassing the multi-frame methods TMP and SSTNet. This may be attributed to the more complex temporal associations in IRDST compared to DAUB. To quantify inter-frame similarity, we employed the mean squared error (MSE) metric to calculate an average MSE of 32.95 for DAUB and 112.03 for IRDST. The higher MSE in IRDST indicates greater complexity in temporal relationships, suggesting that under such conditions, the limited ability of some multi-frame methods to effectively leverage temporal associations may result in their performance being outstripped by single-frame approaches.

Second, our HyperTea achieves the best performance on most evaluation metrics on both datasets, particularly in terms of mAP$_{50}$ and F1 score. For instance, on DAUB, HyperTea attains the highest mAP$_{50}$ of 95.59\% and the highest F1 score of 98.23\%. Furthermore, on IRDST, our HyperTea still achieves the highest mAP$_{50}$ of 76.42\% and F1 score of 87.62\%, far outperforming the previous state-of-the-art (SOTA) method Tridos, which delivers an mAP$_{50}$ of 72.54\% and an F1 score of 85.63\%.

Third, our HyperTea exhibits stronger temporal feature extraction capability, making it more suitable for complex imaging scenarios. As samples in DAUB are relatively stable in the temporal domain, all multi-frame methods perform better on DAUB than on IRDST. For example, the previous multi-frame SOTA method Tridos achieves an mAP$_{50}$ of 91.52\% and an F1 score of 96.17\% on DAUB, but only reaches an mAP$_{50}$ of 72.54\% and an F1 score of 85.63\% on IRDST. Nevertheless, on DAUB, our HyperTea achieves a 4.07\% mAP$_{50}$ improvement and a 2.06\% F1 score improvement over Tridos. On IRDST, the improvement of HyperTea is still significant, with mAP$_{50}$ and F1 score increments over Tridos reaching 3.88\% and 1.99\%, respectively. These comparisons indicate that compared to other methods, HyperTea can more effectively adapt to dynamic temporal scenarios and extract complex temporal features for detection.

\begin{figure*}[h!]
\centering
\includegraphics[width=1\linewidth]{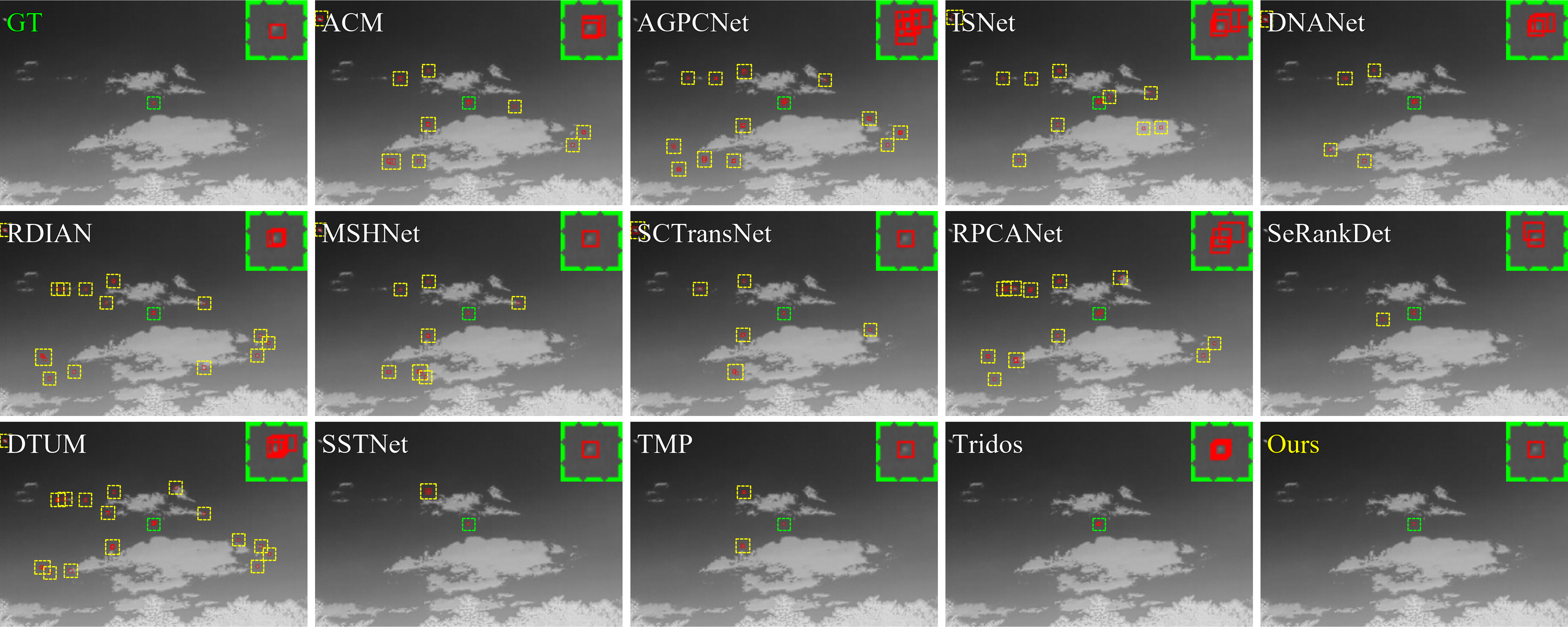}
\caption{Visualization comparisons of 14 methods on IRDST, with 72/266.bmp. GT is ground truth. Red and green boxes represent detected targets and amplified detection regions, respectively. Yellow boxes denote false alarms.} 
\label{fig:visual_methods_1}
\end{figure*}

\begin{figure*}[h!]
\centering
\includegraphics[width=1\linewidth]{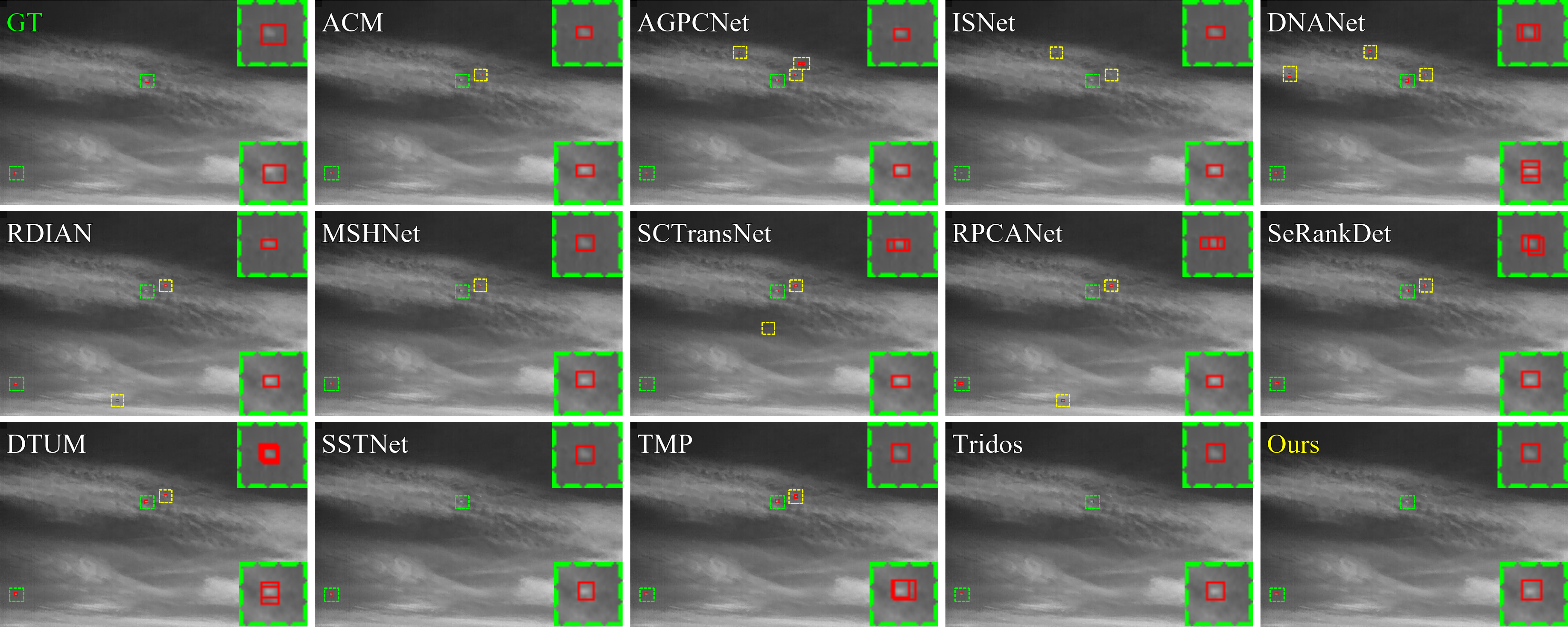}
\caption{Visualization comparisons of 14 methods on IRDST, with 6/14.bmp. GT is ground truth. Red and green boxes represent detected targets and amplified detection regions, respectively. Yellow boxes denote false alarms.} 
\label{fig:visual_methods_2}
\end{figure*}

\begin{figure*}[h!]
\centering
\includegraphics[width=1\linewidth]{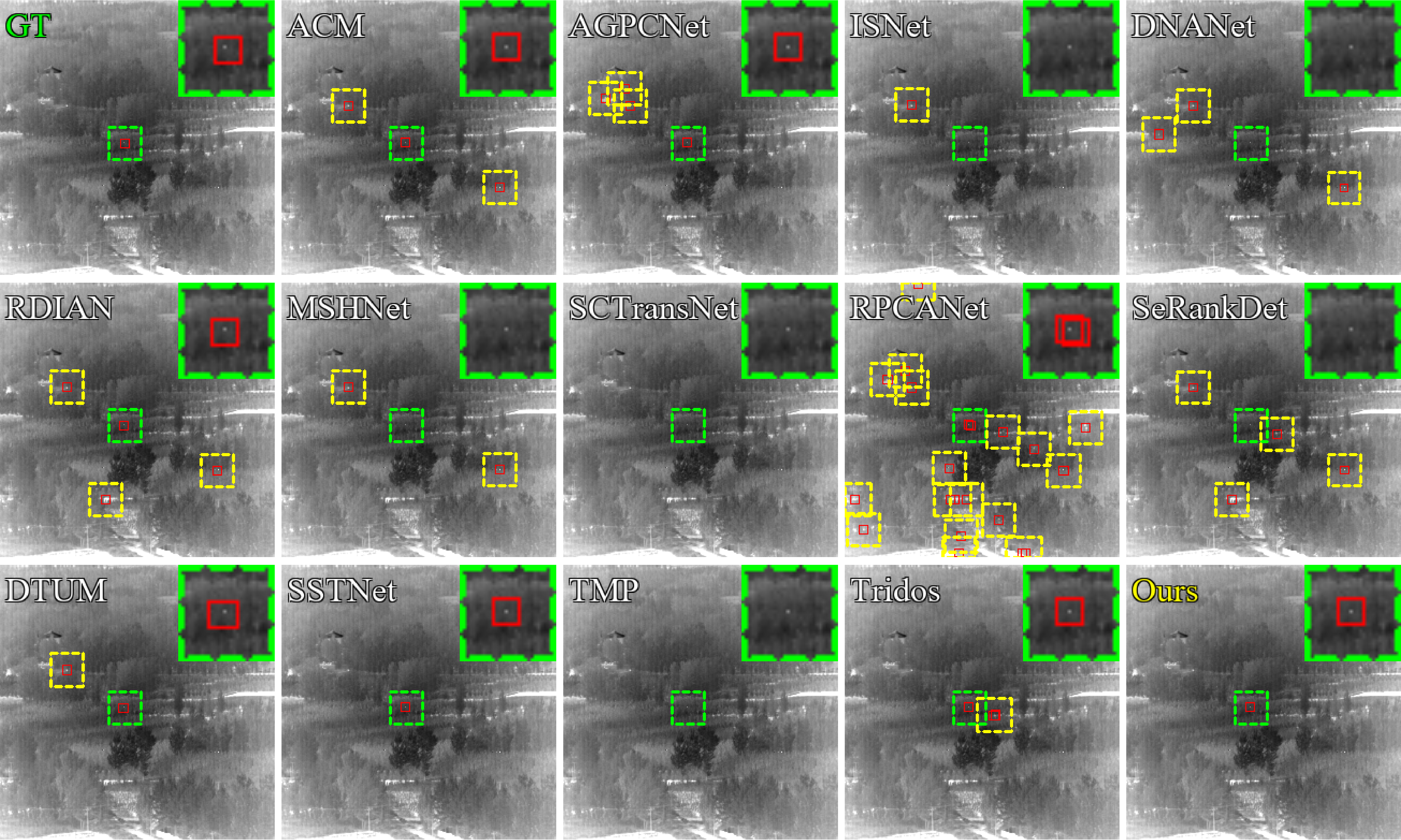}
\caption{Visualization comparisons of 14 methods on DAUB, with 21/433.bmp. GT is ground truth. Red and green boxes represent detected targets and amplified detection regions, respectively. Yellow boxes denote false alarms.} 
\label{fig:visual_methods_3}
\end{figure*}

2) \textit{Visual Comparison:} To provide an intuitive visualization of the detection performances of different methods, we choose three representative scenarios for demonstration, as shown in~\Cref{fig:visual_methods_1,fig:visual_methods_2,fig:visual_methods_3}. It can be observed that our HyperTea consistently achieves accurate detection of moving infrared small targets. Conversely, other methods tend to suffer from missed detections or false alarms.

For example, in~\Cref{fig:visual_methods_1}, on IRDST, our HyperTea can accurately detect the target in complex scenarios with substantial cloud clutter. Conversely, many other methods such as ACM, DNANet, RDIAN, SCTransNet, DTUM, SSTNet, and TMP tend to misclassify clutter as targets.
Furthermore, in~\Cref{fig:visual_methods_2}, when targets are surrounded by cloud clutter, some methods fail to avoid clutter interference and erroneously identify clutter as targets, including DNANet, RDIAN, SCTransNet, DTUM, and TMP.
Meanwhile, as shown in~\Cref{fig:visual_methods_3}, on DAUB, some methods struggle to detect extremely small targets amid complex backgrounds, such as ISNet, MSHNet, SeRankDet, and TMP. Other methods, however, result in false alarms, including ACM, AGPCNet, DTUM, and Tridos.
In summary, the visualization results in these typical scenarios are highly consistent with the quantitative performance presented in~\Cref{table:cmp}, thereby demonstrating the superiority of HyperTea.


\begin{figure}[h!]
\centering
    % 左子图：占单栏宽度的约48%，预留间距
\subfloat{\includegraphics[width=1\linewidth]{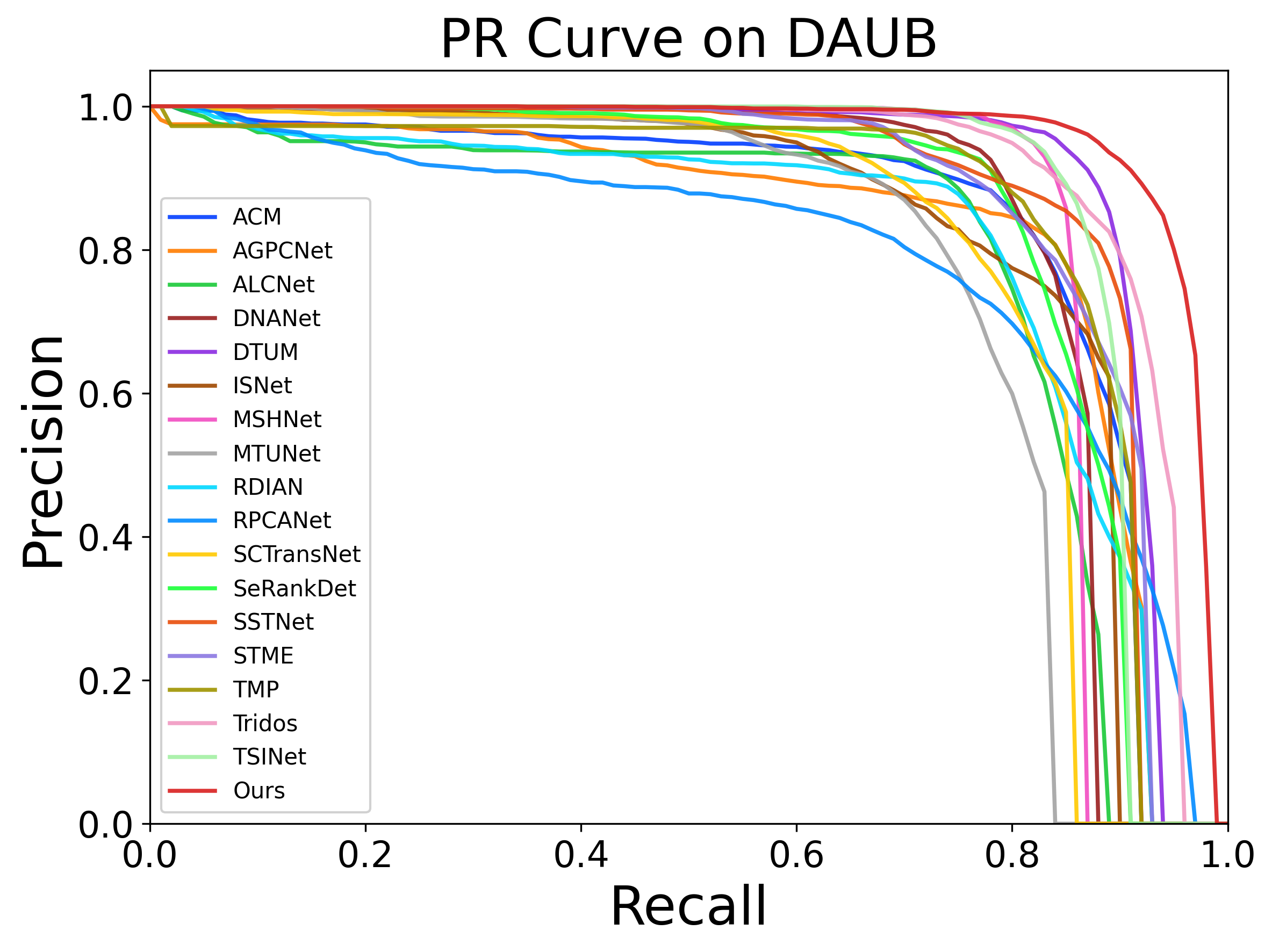}
    }
\hfill  % 填充两图之间的水平间距
    % 右子图：占单栏宽度的约48%
\subfloat{\includegraphics[width=1\linewidth]{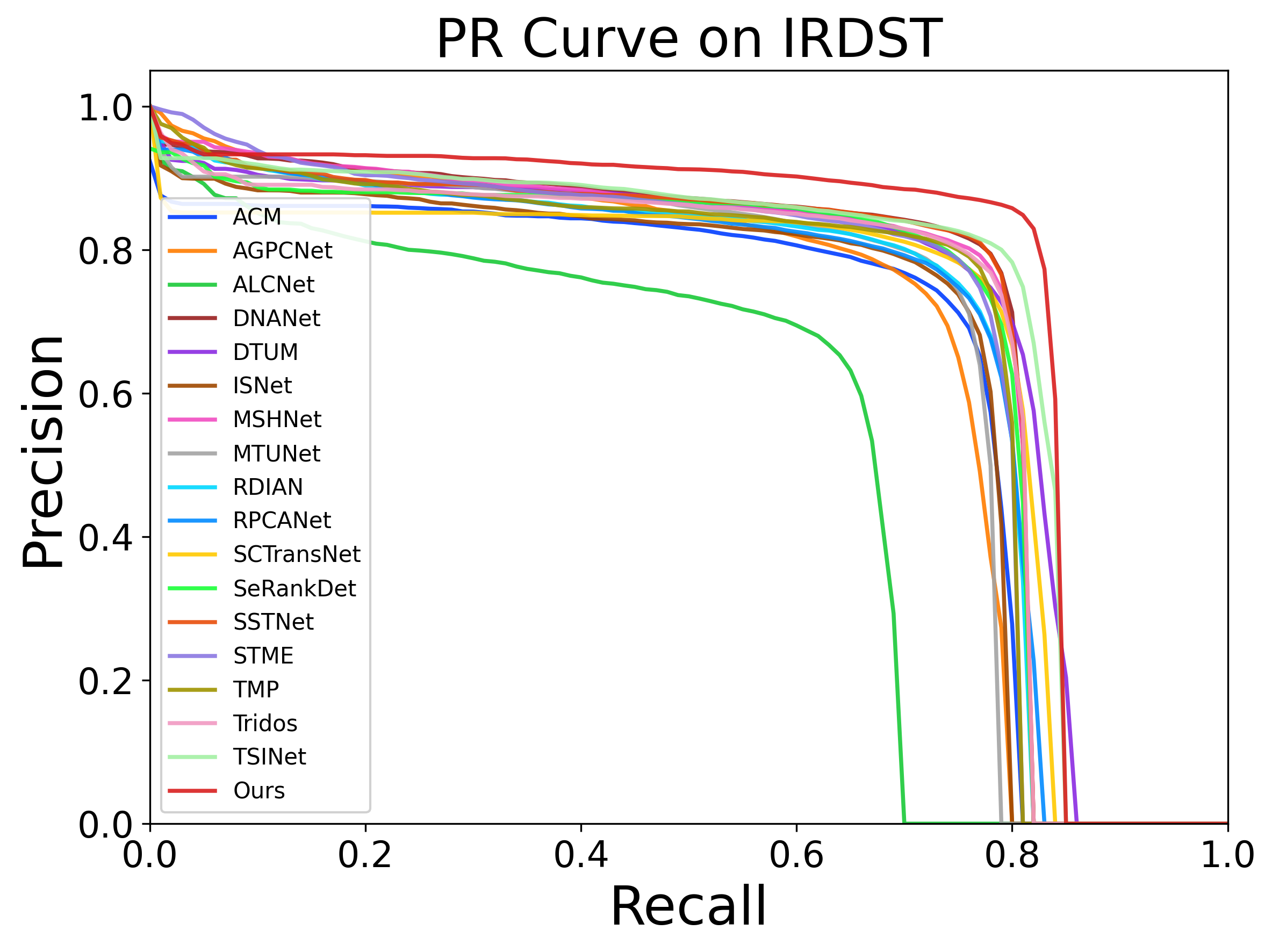}
    }
\caption{PR curves of 18 representative detection methods on datasets DAUB and IRDST.}
\label{fig:pr_curve}
\end{figure}

3) \textit{PR Curve Comparison:} To visually assess the overall performance of different methods, we plot two PR curves on DAUB and IRDST, as shown in~\Cref{fig:pr_curve}. It is readily observable that the PR curve of our HyperTea nearly envelops those of other methods on both datasets. On DAUB, HyperTea's curve consistently dominates the upper-right region relative to competitors; remarkably, it maintains high precision even at relatively high recall. On IRDST, HyperTea's PR curve almost invariably lies above those of other methods. In PR curve analysis, a curve closer to the upper-right corner indicates a superior balance between detection precision and recall. Thus, these two sets of PR curves collectively demonstrate that our HyperTea achieves optimal overall detection performance compared to other methods.

4) \textit{Model Complexity Comparison:}
We compare the model complexity of HyperTea with 15 representative methods in~\Cref{table:complexity_methods}, focusing on Params, GFlops, and frames per second (FPS). Two key observations emerge from the table.

One is that HyperTea exhibits a slight increase in Params and GFlops. Among multi-frame methods, DTUM has the smallest params at 2.79M, whereas HyperTea has a moderate 13.75M, lower than TMP (16.41M) and Tridos (14.13M). In terms of GFlops, TMP achieves the best result among multi-frame methods at 92.85\%, whereas HyperTea registers 148.66. A probable reason is that TMP is based on a single temporal scale, while HyperTea learns feature representations across both local and global temporal scales and requires cross-scale alignment, which entails additional computations. Nevertheless, these complexity costs are acceptable and worthwhile given HyperTea's significant performance advantages.

The other observation is that HyperTea achieves an excellent balance between performance and FPS. As multi-frame methods are based on spatiotemporal domains, they typically outperform single-frame methods in performance but often have lower FPS. For instance, DNANet, the SOTA single-frame method, achieves an F1 score of 84.86\%---roughly on par with SSTNet's 84.76\%, but far below HyperTea's 87.62\%. Additionally, its FPS 12.13 is also lower than HyperTea's 14.52. Moreover, among multi-frame methods, HyperTea boasts the highest FPS, even surpassing some single-frame methods such as DNANet and SCTransNet. This underscores its superior balance between detection performance and inference efficiency.

\begin{table}[h!]
\centering
\captionsetup{justification=centering} % 设置标题居中

\resizebox{\linewidth}{!}{ 
\begin{tabular}{c|c|c|cc|ccc}
\hline
\textbf{Methods} & 
\textbf{Arch} & 
\textbf{Frames} & 
\textbf{mAP$_{50}\uparrow$} & 
\textbf{F1$\uparrow$} &
\textbf{Params$\downarrow$} &
\textbf{GFlops$\downarrow$} &
\textbf{FPS$\uparrow$} \\
\hline
% --- Single-frame ---
ACM~\cite{daiAsymmetricContextualModulation2021} & CNN & 1 & 65.31 & 81.40 & 2.89M & 24.14 & 51.55 \\ 
ALC~\cite{daiAttentionalLocalContrast2021} & CNN & 1 & 53.10 & 72.93 & 2.92M & \textbf{23.95} & \textbf{56.23} \\ 
ISNet~\cite{zhangISNetShapeMatters2022} & CNN & 1 & 66.30 & 81.96 & 3.46M & 265.87 & 26.82 \\ 
AGPCNet~\cite{zhangAttentionGuidedPyramidContext2023b} & CNN & 1 & 67.10 & 82.47 & 14.85M & 366.37 & 15.72 \\ 
DNANet~\cite{liDenseNestedAttention2023} & CNN & 1 & 71.39 & 84.86 & 7.19M & 135.00 & 12.13 \\ 
RDIANet~\cite{sunReceptiveFieldDirectionInduced2023} & CNN & 1 & 68.70 & 83.54 & \textbf{2.71M} & 50.67 & 45.64 \\ 
MSHNet~\cite{liuInfraredSmallTarget2024} & CNN & 1 & 71.04 & 84.80 & 6.55M & 69.78 & 25.73 \\ 
RPCANet~\cite{wuRPCANetDeepUnfolding} & CNN & 1 & 68.80 & 83.33 & 3.17M & 377.48 & 26.40 \\ 
MTUNet~\cite{wuMTUNetMultilevelTransUNet2023} & CNN+Trans & 1 & 67.20 & 82.35 & 9.13M & 67.42 & 48.79 \\ 
SCTransNet~\cite{yuanSCTransNetSpatialChannelCross2024c} & CNN+Trans & 1 & 68.41 & 83.01 & 35.74M & 147.02 & 14.22 \\ 
SeRankDet~\cite{daiPickBunchDetecting2024a} & CNN+Trans & 1 & 69.50 & 83.70 & 111.37M & 1147 & 18.54 \\ 
% --- Multi-frame ---
DTUM~\cite{liDirectionCodedTemporalUShape2023a} & CNN & 5 & 71.80 & 85.24 & 2.79M & 103.73 & 12.85 \\ 
STME~\cite{pengMovingInfraredDim2025a} & CNN & 5 & 69.95 & 87.64 & 9.85M & 41.92 & 15.26 \\ 
SSTNet~\cite{chenSSTNetSlicedSpatioTemporal2024} & CNN+RNN & 5 & 70.91 & 84.76 & 11.95M & 123.60 & 12.97 \\ 
TMP~\cite{zhuTMPTemporalMotion2024c} & CNN+Trans & 5 & 69.12 & 83.68 & 16.41M & 92.85 & 14.02 \\ 
Tridos~\cite{duanTripleDomainFeatureLearning2024a} & CNN+Trans & 5 & 72.54 & 85.63 & 14.13M & 130.72 & 13.24 \\ 
TSINet~\cite{zhuangTemporalSemanticInteractionNetwork2025b} & CNN+Trans & 5 & 73.19 & 86.19 & 8.79M & 210.55 & 12.68 \\ 
HyperTea(Ours) & CNN+RNN+HGNN & 5 & \textbf{76.42} & \textbf{87.62} & 13.75M & 148.66 & 14.52 \\
\hline
\end{tabular}
}
\caption{Complexity comparisons of inference on IRDST.}
\label{table:complexity_methods}
\end{table}

\subsection{Ablation Study}
1) \textit{Effects of Different Components:}
To investigate the effectiveness of each component in HyperTea, we conducted a series of ablation studies on DAUB and IRDST, as shown in~\Cref{table:aba_components}. For the ablation experiments of TAM, we use 1×1 convolutions to adjust the number of channels, aiming to verify the module's effectiveness while minimizing changes to the network structure.

Two key findings can be drawn from~\Cref{table:aba_components}. First, each component contributes to improving detection performance to some extent, and the best performance is achieved when all components are applied together. For example, on IRDST, the baseline without any components only achieves an mAP$_{50}$ of 66.80\% and an F1 score of 81.94\%. After applying LTEM alone, the mAP$_{50}$ increases to 69.42\% and the F1 score to 83.55\%. When GTEM is applied alone, the mAP$_{50}$ and F1 score are further improved to 71.44\% and 84.99\%, respectively. Additionally, the performance of TAM varies with different inputs. When features enhanced by both LTEM and GTEM are fed into TAM, the mAP$_{50}$ and F1 score on DAUB increase to 95.59\% and 98.23\%, respectively; on IRDST, they reach 76.42\% and 87.62\%, achieving the best performance.
Second, simple 1×1 convolutions to realize cross-temporal feature fusion effectively. For instance, on IRDST, the model without TAM only achieves an mAP$_{50}$ of 69.07\% and an F1 score of 83.25\%, which is lower than those of the model with LTEM alone (mAP$_{50}$: 69.42\%, F1: 83.55\%) and the model with GTEM alone (mAP$_{50}$: 71.44\%, F1: 84.99\%).

\begin{table*}[h!]
\centering
\captionsetup{justification=centering} % 设置标题居中

\resizebox{\linewidth}{!}{ 
\begin{tabular}{c|c|c|c|cccc|cccc}
% \begin{tabular}{|ccccccccccc}
\hline
\multirow{2}{*}{\textbf{Settings}} & \multirow{2}{*}{LTEM} & \multirow{2}{*}{GTEM} & \multirow{2}{*}{TAM} & 
\multicolumn{4}{c}{IRDST} &
\multicolumn{4}{c}{DAUB}\\ \cline{5-12}
&&&&\textbf{mAP$_{50}$} & Pr & Re & F1 
&\textbf{mAP$_{50}$} & Pr & Re & F1 \\ \hline
w/o All  & - & - & - 
& 66.80 & 84.95 & 79.13 & 81.94
&82.63&95.51&87.32&91.23\\ 
w/ LTEM  & \checkmark & - & - 
& 69.42 & 88.01 & 79.52 & 83.55
&89.05&94.47&95.95&95.21\\  
w/ GTEM  & - & \checkmark & - 
& 71.44 & 88.28 & 81.93 & 84.99
&89.79&96.97&93.82&95.37\\  
w/ LTEM \& GTEM  & \checkmark & \checkmark & - 
& 69.07 & 86.67  & 80.09 & 83.25
&89.74&97.84&92.67&95.19\\  
w/ TAM(A)  & \checkmark & - & \checkmark 
& 72.18 & 89.25  & 81.66 & 85.28 
&90.92&96.18&95.45&95.81\\  
w/ TAM(B)  & - & \checkmark & \checkmark 
& 73.27 & 89.49  & 82.52 & 85.86 
&91.05&95.53&96.66&96.09\\  
w/All  & \checkmark & \checkmark & \checkmark 
&\textbf{76.42}&\textbf{91.18}
&\textbf{84.33}&\textbf{87.62} 
& \textbf{95.59} & \textbf{98.14} & \textbf{98.31} &\textbf{98.23}\\ 
\hline
\end{tabular}
}
\caption{Ablation study on different components. TAM(A) uses $\mathbf{L}_{st}$ as query and $\mathbf{F}_s$ as key \& value. TAM(B) uses $\mathbf{F}_T$ as query and $\mathbf{G}_{st}$ as key \& value.}
\label{table:aba_components}
\end{table*}

2) \textit{Effects of DPCB and HCU in GTEM:}
To explore the potential contributions of HCU and DPCB to GTEM, we conducted a set of ablation experiments, as shown in~\Cref{table:aba_GTEM}. First, DPCB provides more robust feature representations by fusing multi-directional features, thereby improving detection performance. For example, with DPCB, on DAUB, mAP$_{50}$ and F1 can be increased from 91.06\% to 92.52\% and from 95.91\% to 96.68\%, respectively.  Similarly, on IRDST, HCU can boost mAP$_{50}$ from 69.67\% to 71.84\% and F1 from 83.75\% to 85.11\%. These effects can be attributed to HCU's strong capability in modeling high-order correlations. Notably, removing the $1\times5$ and $5\times1$ convolutions (w/o H\&V Conv) leads to a noticeable performance drop (e.g., a 3.55\% decrease in $mAP_{50}$ on IRDST), which validates the necessity of our tailored convolution.

3) \textit{Effects of GLTA and CSAM in TAM:}
Furthermore, we investigate the impacts of GLTA and CSAM in~\Cref{table:aba_TAM}, from which three distinct findings can be derived. First, GLTA effectively aligns the features enhanced by LTEM and GTEM. For instance, on DAUB, with GLTA, mAP$_{50}$ and F1 are improved from 89.74\% to 92.13\% and from 95.19\% to 96.61\%, respectively. Second, CSAM can further enhance the aligned features. On IRDST, for example, using CSAM boosts mAP$_{50}$ from 73.30\% to 76.42\% and F1 from 85.85\% to 87.62\%. Third, the performance gain from using CSAM alone is limited. On DAUB, for instance, with CSAM alone merely increases mAP$_{50}$ from 89.74\% to 89.81\% and F1 from 95.19\% to 95.23\%. This phenomenon also reflects the severe impact of misalignment between features of different temporal scales on detection performance.

% \begin{center}
% \begin{minipage}{\linewidth}
\begin{table}[h!]
\centering
\captionsetup{justification=centering} % 设置标题居中

\resizebox{\linewidth}{!}{ 
\begin{tabular}{c|cccc|cccc}
\hline
\multirow{2}{*}{\textbf{Settings}} & 
\multicolumn{4}{c}{IRDST} &
\multicolumn{4}{c}{DAUB}\\ \cline{2-9}
& mAP$_{50}$ & Pr & Re & F1  
& mAP$_{50}$ & Pr & Re & F1  \\ \hline
GTEM w/o All 
& 69.67 & 87.15 & 80.61 & 83.75
&91.06&97.22&94.64&95.91\\
GTEM w HCU
& 71.84 & 88.76 & 81.76 & 85.11
&92.33&97.59&95.55&96.56\\
GTEM w DPCB 
& 71.99 & 89.07 & 81.78 & 85.27
&92.52&96.13&97.24&96.68\\ 
GTEM w/o H\&V Conv 
&72.87&89.01&82.63&85.69
&93.45&97.01&97.97&97.48\\ 
GTEM 
&\textbf{76.42}&\textbf{91.18}
&\textbf{84.33}&\textbf{87.62} 
& \textbf{95.59} & \textbf{98.14} & \textbf{98.31} &\textbf{98.23}\\ 
\hline
\end{tabular}
}
\captionof{table}{Ablation study on DPCB and HCU of GTEM.}
\label{table:aba_GTEM}

\end{table}
% \end{minipage}
% \hfill
% \begin{minipage}{\linewidth} % 第一个表格占 45% 的宽度

\begin{table}[h!]
\centering
\captionsetup{justification=centering} % 设置标题居中

\resizebox{\linewidth}{!}{
\begin{tabular}{c|cccc|cccc}
\hline
\multirow{2}{*}{\textbf{Settings}} &
\multicolumn{4}{c}{IRDST} &
\multicolumn{4}{c}{DAUB} \\ \cline{2-9}
& mAP$_{50}$ & Pr & Re & F1  
& mAP$_{50}$ & Pr & Re & F1 \\ \hline
TAM w/o All   
& 69.07 & 86.67  & 80.09 & 83.25 
&89.74&97.84&92.67&95.19\\  
TAM w GLTA   
& 73.30 & 88.76 & 83.13 & 85.85
&92.13&95.73&97.52&96.61\\  
TAM w CSAM
& 69.78 & 87.36 & 80.81 & 83.96
&89.81&95.08&95.39&95.23\\ 
TAM 
& \textbf{76.42} & \textbf{91.16} & \textbf{84.28} & \textbf{87.62}
& \textbf{95.59} & \textbf{98.14} & \textbf{98.31} &\textbf{98.23}\\ 
\hline
\end{tabular}
}
\captionof{table}{Ablation study on GLTA and CSAM of TAM.} % 表格标题
\label{table:aba_TAM}

\end{table}
% \end{minipage}
% \end{center}
4) \textit{Effects of layer number and patch size in LTEM:}
As shown in~\Cref{table:aba_LTEM_l_ps}, we designed six sets of experiments to evaluate the impact of layer number (L) and patch size (ps) on LTEM. From these experiments, we can draw the following three observations. First, when L$=$1 and ps$=$2, the mAP$_{50}$ on DAUB and IRDST can reach peak values of 95.59\% and 76.42\%, respectively. In the experiments, any other settings generally fail to enable HyperTea to achieve the optimal performance. Second, when the number of layers increases, HyperTea exhibits a significant performance degradation on both DAUB and IRDST. For example, on DAUB, when L increases from 1 to 2, mAP$_{50}$ decreases from 95.59\% to 88.07\%. A possible reason for this observation is that HyperTea suffers from overfitting when the number of layers is greater than 1. Third, performance degrades when ps is extremely small or large. For instance, on DAUB, L$=$1, when ps decreases from 2 to 1, mAP$_{50}$ drops from 95.59\% to 92.31\%; when ps increases from 2 to 4, mAP$_{50}$ decreases from 95.59\% to 93.04\%. A possible reason for this observation is that when ps$=$1, the semantic features of a single pixel are insufficient to effectively model long-range dependencies; when ps$=$4, an excessively large size leads to the loss of low-level details. Only when ps$=$2 can the optimal performance be achieved by enriching feature representations while well preserving low-level details.

\begin{table}[h!]
\centering
\captionsetup{justification=centering} % 设置标题居中

\resizebox{\linewidth}{!}{ 
\begin{tabular}{c|c|cccc|cccc}
\hline 
\multirow{2}{*}{$L$} & \multirow{2}{*}{$ps$} & 
\multicolumn{4}{c}{IRDST} &
\multicolumn{4}{c}{DAUB} \\ \cline{3-10}
&& 
\textbf{mAP$_{50}$} & Pr & Re & F1&
\textbf{mAP$_{50}$} & Pr & Re & F1 \\ \hline
1 & 1
& 71.83 & 88.78 & 81.74 & 85.11
&92.31&95.90&97.64&96.76\\ 
1 & 2
&\textbf{76.42}&\textbf{91.18}
&\textbf{84.33}&\textbf{87.62} 
& \textbf{95.59} & \textbf{98.14} & \textbf{98.31} &\textbf{98.23}\\ 
1 & 4
& 69.80 & 88.54 & 79.66 & 83.87
&93.04&96.67&97.22&96.94\\  
2 & 1
& 70.27 & 87.96 & 80.84 & 84.25
&89.68&93.22&97.41&95.27\\ 
2 & 2
& 71.69 & 88.70 & 81.49 & 84.94
&88.07&92.47&96.28&94.34\\  
2 & 4
& 71.06 & 89.10 & 80.20 & 84.41
&92.34&95.95&97.83&96.88\\  
\hline
\end{tabular}
}
\captionof{table}{With different layer number $L$ and patch size $ps$ of LTEM.}
\label{table:aba_LTEM_l_ps}
\end{table}
% \end{minipage} 
% \hfill
% \begin{minipage}{\linewidth}
% \begin{minipage}[t]{0.3\textwidth}

\begin{table}[h!]
\centering
\captionsetup{justification=centering} % 设置标题居中

\resizebox{\linewidth}{!}{ 
\begin{tabular}{c|cccc|cccc}
\hline
\multirow{2}{*}{\textbf{Settings}} & 
\multicolumn{4}{c}{IRDST} &
\multicolumn{4}{c}{DAUB}\\ \cline{2-9}
& \textbf{mAP$_{50}$} & Pr & Re & F1 
& \textbf{mAP$_{50}$} & Pr & Re & F1 \\ \hline
6 
& 72.14 & 88.36 & 82.24 & 85.19
&93.25&95.99&98.22&97.09\\ 
7
& 72.96 & 89.01 & 82.85 &85.82 
&92.86&96.36&97.93&97.14 \\  
8 
&\textbf{76.42}&\textbf{91.18}
&\textbf{84.33}&\textbf{87.62} 
& \textbf{95.59} & \textbf{98.14} & \textbf{98.31} &\textbf{98.23}\\
9
& 71.15 & 87.99 & 81.66 &84.71 
&88.54&92.09&97.62&94.78\\ 
10
& 70.31 & 87.04 & 81.32 &84.08 
&92.33&94.99&98.22&96.58\\ 
\hline
\end{tabular}
}
\captionof{table}{With different hypergraph distance threshold.}
\label{table:aba_hg_thre}
\end{table}
% \end{minipage}
% \end{center}

5) \textit{Effects of Hypergraph distance threshold:}
We further conducted ablation experiments to investigate the impact of the distance threshold used in hypergraph construction, with the results listed in~\Cref{table:aba_hg_thre}. It can be observed that when the threshold is 8, the mAP$_{50}$ on DAUB and IRDST reaches peak values of 95.59\% and 76.42\% respectively, and both excessively low and high thresholds lead to performance degradation. This is because a higher threshold results in more densely connected hypergraphs, where noisy nodes are mixed into hyperedges, potentially causing features to lack discriminability. However, a lower threshold leads to sparser hypergraphs, preventing the high-order learning capability of the hypergraph from being fully utilized. Therefore, our HyperTea is constructed using a distance threshold of 8.

\begin{figure}[t]
\centering
    \subfloat{\includegraphics[width=.48\linewidth]{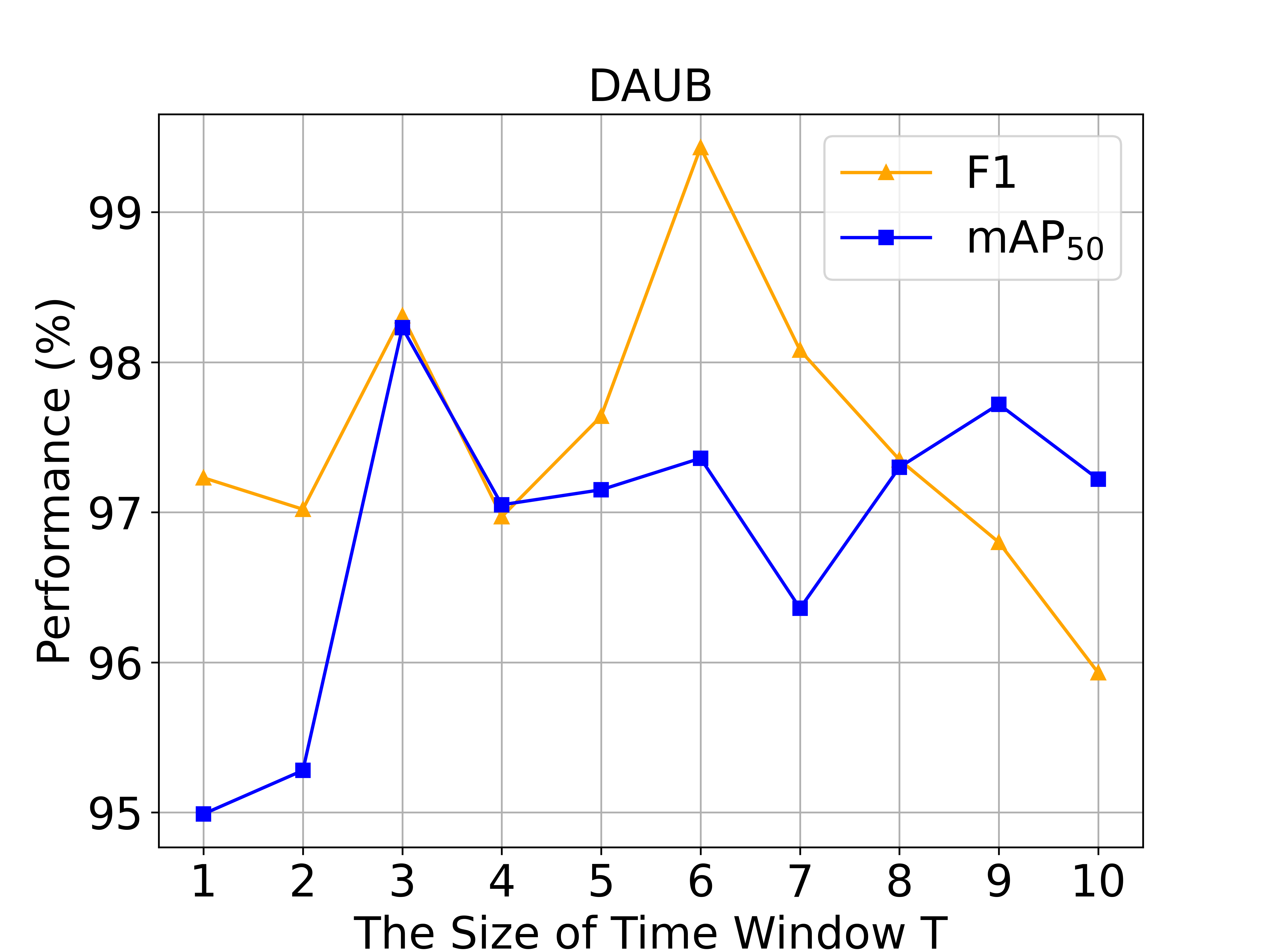}
    }
\hfil % 自动填充水平间距
    \subfloat{\includegraphics[width=.48\linewidth]{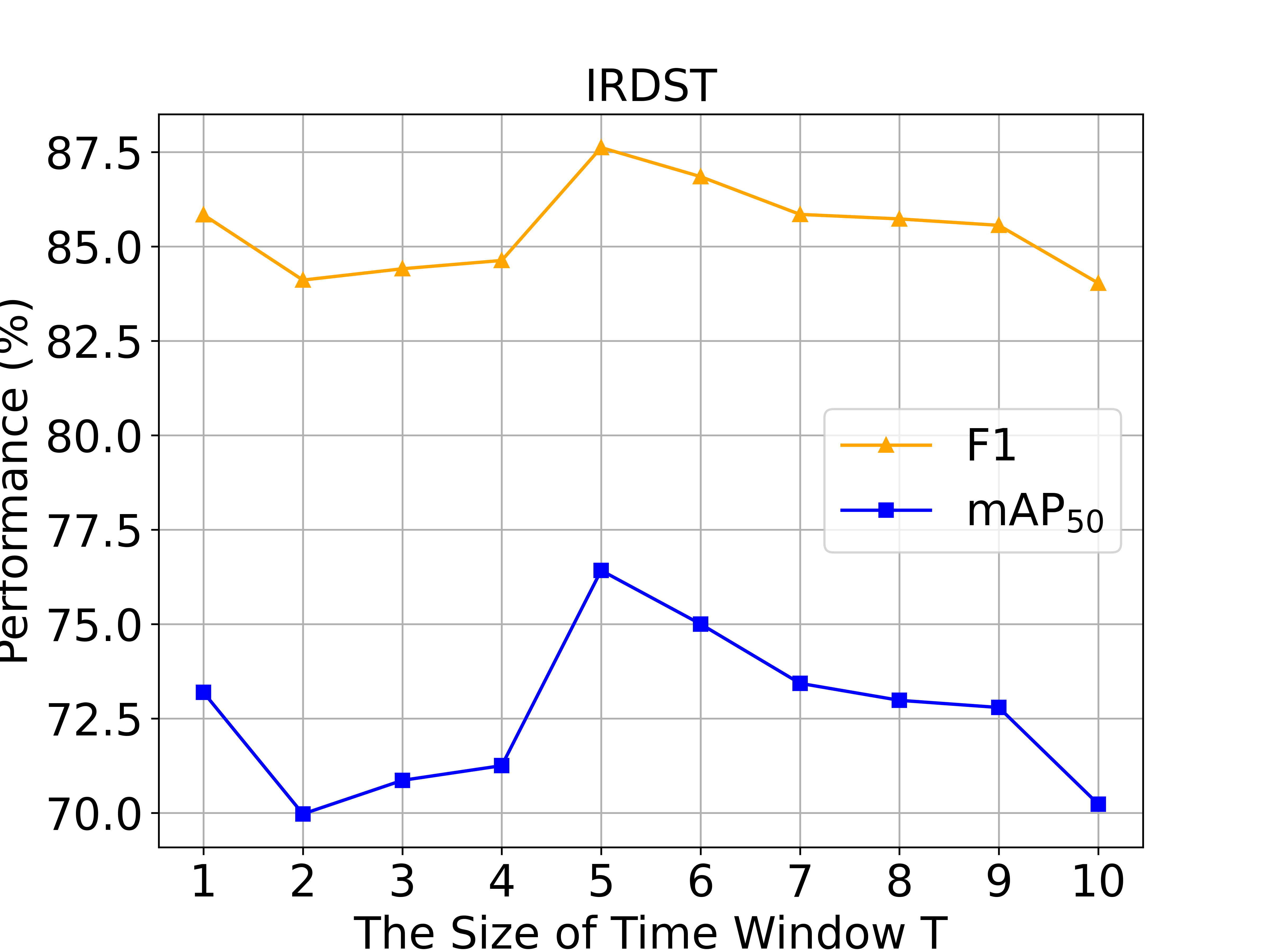}
    }
\caption{Effects of time window size T on HyperTea.}
\label{fig:time_window}

\end{figure}

6) \textit{Effects of Time Window Size T:} 
We conduct a group of experiments with different time window sizes to investigate the impact of time window T on detection performance, as shown in~\Cref{fig:time_window}. 

First, it can be observed that the impact of T on detection performance differs between DAUB and IRDST. On DAUB, HyperTea achieves peak mAP$_{50}$ and F1 values when T$=$3. On IRDST, however, the peak mAP$_{50}$ and F1 values are reached when T$=$5. The discrepancy in the T values corresponding to these peaks is mainly caused by differences between the datasets. As mentioned earlier, the average MSE values of DAUB and IRDST are 32.95 and 112.03 respectively, meaning the temporal dynamics of IRDST are more complex, making the learning of its temporal features more challenging and thus requiring more frames. In contrast, the temporal dynamics of DAUB are relatively stable, therefore,  the temporal context provided by 3 frames is sufficient for detection.

Additionally, an appropriate T is crucial for IRSTD. For example, on DAUB, when T $\leq$ 3, increasing T leads to a noticeable improvement in performance. When T $>$ 3, the detection performance first decreases and then fluctuates, showing an overall tendency to stabilize. This is consistent with the temporal dynamic characteristics of DAUB: with smaller MSE, 3 frames are sufficient for detection, longer temporal context provides limited gains and may even risk performance degradation owing to feature redundancy and clutter interference. 

7) \textit{Comparison With Hyper-YOLO:} 
To further verify the advantages of our HyperTea, we compared it with Hyper-YOLO, as shown in~\Cref{table:cmp_hyperyolo}. It can be observed that our HyperTea has a significant leading advantage over Hyper-YOLO on both datasets. For instance, on DAUB, HyperTea attains an mAP$_{50}$ of 95.59\%, Precision (Pr) of 98.14\%, Recall (Re) of 98.31\%, and F1 of 98.23\%, which are far superior to those of Hyper-YOLO. This set of quantitative comparisons indicates that the main advantage of our HyperTea lies in fully leveraging the complementary advantages between basic network structures and performing feature enhancement and alignment across different temporal scales, rather than simply integrating the hypergraph module into the detection framework.

% \begin{minipage}{\linewidth} % 第一个表格占 45% 的宽度
\begin{table}[h!]
\centering
\captionsetup{justification=centering} % 设置标题居中

\resizebox{\linewidth}{!}{
\begin{tabular}{c|cccc|cccc}
\hline 
\multirow{2}{*}{Methods} & 
\multicolumn{4}{c}{IRDST} &
\multicolumn{4}{c}{DAUB}\\ \cline{2-9}
&\textbf{mAP$_{50}$} & Pr & Re & F1 
&\textbf{mAP$_{50}$} & Pr & Re & F1 \\ \hline
Hyper-YOLO
&69.90&86.67&81.59&84.05
&88.44&96.45&92.74&94.56\\ 
Ours
&\textbf{76.42}&\textbf{91.18}
&\textbf{84.33}&\textbf{87.62}
& \textbf{95.59} & \textbf{98.14} & \textbf{98.31} &\textbf{98.23}\\ 
\hline
\end{tabular}
}
\captionof{table}{Performance Comparison With Hyper-YOLO.} % 表格标
\label{table:cmp_hyperyolo}

\end{table}
% \end{minipage}

\subsection{Discussion}
The effectiveness and superiority of our HyperTea detection model have been validated across two infrared sequence datasets, highlighting its outstanding performance across diverse scenarios. While our method follows the common paradigm of extracting and fusing multiscale spatio-temporal features, it possesses several distinctive advantages. 
Specifically, owing to the multi-timescale modeling, our model can simultaneously extract temporal cues at both local and global scales, whereas conventional models are typically restricted to a single temporal scale. 
Furthermore, to the best of our knowledge, this is the first approach toward incorporate hypergraph modeling to enhance the representation of global high-order correlations, addressing the inherent limitations of existing models that focus on local low-order relationships, such as local receptive field optimizations in CNNs or pair-wise associations in Transformers. Consequently, our approach demonstrates substantial advantages in feature extraction and fusion compared to other state-of-the-art models.

Nevertheless, we acknowledge several limitations within the current version of HyperTea. 
First, a primary concern is the spatio-temporal complexity. Although the integration of CNN, RNN, and HGNN provides powerful representation capabilities, it inevitably introduces higher computational overhead compared to single-architecture models. 
The current complexity and processing speed remain at a moderate level, which may constrain its deployment in real-time or resource-limited environments. 
Second, the current tri-architecture fusion paradigm is not necessarily the optimal configuration. 
While it transcends the constraints of individual conventional paradigms, the structural synergy between CNN, RNN, and HGNN still offers substantial room for optimization to achieve a better balance between performance and efficiency. 
Third, the distance threshold in hypergraph construction remains a sensitive hyperparameter. 
Currently, this threshold should be meticulously determined through theoretical prior and experiments to obtain optimal values across different scenarios and data distributions, as a fully adaptive selection mechanism is yet to be developed. 
Additionally, while our HyperTea elevates primary performance metrics, a slight reduction in recall may occasionally occur under extreme dynamic conditions, indicating a risk of missing true positives in special scenarios not encountered during training. 

From a practical perspective, the ablation results and the above discussion also suggest several directions for improving the applicability of our HyperTea. 
For instance, lightweight design strategies could be explored to reduce computational overhead, such as simplifying the hypergraph computation unit or exploring a unified framework for multi-timescale context modeling. 
In addition, real-time inference can be further improved by adopting model compression techniques, including pruning, quantization, and knowledge distillation. 
These strategies may help achieve a better balance between detection performance and computational efficiency, making our HyperTea more suitable for real-world deployment.

\section{Conclusion}
\label{sec:conclusion}
To achieve performance improvement by cross-temporal representation learning of moving infrared dim-small targets, we propose HyperTea, a network architecture combining CNNs, RNNs, and HGNNs. We first design a GTEM to enhance the global temporal context. Then, we develop the LTEM to enhance the local temporal context in which key frames are located. Furthermore, we utilize HGNNs to endow GTEM and LTEM with the capability of high-order learning. To bridge the potential feature misalignment, we design the TAM to facilitate the fusion of features across different temporal scales. Comparative experiments on DAUB and IRDST demonstrate the distinct superiority of our HyperTea. It outperforms existing state-of-the-art (SOTA) methods on most metrics, with moderate computational complexity and FPS. Ablation studies further indicate that all designed components contribute significantly to improving the detection performance. In experiments, we also find that the temporal similarity of datasets (measured by MSE) critically affects detection performance, which should provide insights for the design of sampling strategies and modules using multiframe inputs. In the future, designing more effective temporal feature sampling, capture, and enhancement schemes can be explored.

\section*{CRediT authorship contribution statement}
\textbf{Zhaoyuan Qi:} Conceptualization, Methodology, Writing – Original draft preparation, Data curation, Investigation, Visualization.
\textbf{Weihua Gao:} Data curation, Formal analysis, Investigation, Writing – Reviewing and Editing.
\textbf{Wenlong Niu:} Supervision, Funding acquisition, Writing – Reviewing and Editing.
\textbf{Jie Tang:} Software, Validation, Resources.
\textbf{Xiaodong Peng:} Project administration, Resources.

% 利益冲突声明
\section*{Declaration of Competing Interest}
The authors declare that they have no known competing financial interests or personal relationships that could have appeared to influence the work reported in this paper.

\section*{Data availability} 
The data used in this study are publicly available.

\section*{Acknowledgments}
This work was supported in part by the Youth Innovation Promotion Association under Grant E1213A02; and in part by the Key Research Program of Frontier Sciences, Chinese Academy of Sciences (CAS), under Grant 22E0223301.
{
    % \smalls
    \bibliographystyle{elsarticle-num}
    \bibliography{ref}
}
\end{document}